\documentclass[lettersize,journal]{IEEEtran}

\usepackage{amsmath,amsfonts}
\usepackage{algorithmic}
\usepackage{algorithm}
\usepackage{array}
\usepackage[caption=false,font=normalsize,labelfont=sf,textfont=sf]{subfig}
\usepackage{textcomp}
\usepackage{stfloats}
\usepackage{url}
\usepackage{verbatim}
\usepackage{graphicx}
\usepackage{cite}

\usepackage{xcolor}
\usepackage[
    colorlinks=true,
    linkcolor=black,
    citecolor=blue,
    urlcolor=blue
]{hyperref}
\hypersetup{colorlinks=false}

\usepackage{bm}
\usepackage{multirow}

\makeatletter
\renewcommand\fs@ruled{%
  \def\@fs@cfont{\bfseries\color{black}}%
  \let\@fs@capt\floatc@ruled
  \def\@fs@pre{\color{black}\hrule height.8pt depth0pt \kern2pt}%
  \def\@fs@post{\kern2pt\color{black}\hrule\relax}%
  \def\@fs@mid{\kern2pt\color{black}\hrule\kern2pt}%
  \let\@fs@iftopcapt\iftrue
}
\makeatother

\begin{document}

\title{DifAttack++: Query-Efficient Black-Box Adversarial Attack via Hierarchical Disentangled Feature Space in Cross--Domain}

\author{Jun~Liu,~\IEEEmembership{Member,~IEEE},
Jiantao~Zhou,~\IEEEmembership{Senior~Member,~IEEE}, Jiandian~Zeng,~\IEEEmembership{Member,~IEEE}, Jinyu~Tian,~\IEEEmembership{Member,~IEEE}, and Isao Echizen,~\IEEEmembership{Senior~Member,~IEEE}
    \thanks{
J. Liu and J. Zhou are with the State Key Laboratory of Internet of Things for Smart City and the Department of Computer and Information Science, Faculty of Science and Technology, University of Macau, Macau 999078, China (e-mail:yc07453@connect.umac.mo;jtzhou@umac.mo). J. Liu is also with the Echizen
Laboratory, National Institute of Informatics, Tokyo 101-8430, Japan. \emph{Corresponding author: Jiantao Zhou.} 

J. Zeng is with the Institute of Artificial Intelligence and Future Networks, Beijing Normal University, Zhuhai 519087, China (email: jiandian@bnu.edu.cn).

J. Tian is with the School of Computer Science and Engineering, Macau University of Science and Technology, Macau 999078, China (e-mail: jytian@must.edu.mo).

I. Echizen is with the Echizen
Laboratory, National Institute of Informatics, and University of Tokyo, Tokyo 101-8430, Japan
(e-mail: iechizen@nii.ac.jp).
}
}

\markboth{}%
{Shell \MakeLowercase{\textit{et al.}}: A Sample Article Using IEEEtran.cls for IEEE Journals}


\maketitle

\begin{abstract}
This work investigates efficient score-based black-box adversarial attacks that achieve a high Attack Success Rate ({ASR}) and good generalization ability. We propose a novel attack framework, termed {DifAttack++}, which operates in a hierarchical disentangled feature space and significantly differs from existing methods that manipulate the entire feature space. Specifically, DifAttack++ firstly disentangles an image’s latent representation into an {Adversarial Feature} (\textbf{AF}) and a \textbf{Visual Feature} (\textbf{VF}) using an autoencoder equipped with a carefully designed \textbf{Hierarchical Decouple-Fusion} (\textbf{HDF}) module. In this formulation, the AF primarily governs the adversarial capability of an image, while the VF largely preserves its visual appearance. To enable the feature disentanglement and image reconstruction, we jointly train two autoencoders for the clean and adversarial image domains, i.e., cross-domain, respectively, using paired clean images and their corresponding \textbf{Adversarial Examples} (\textbf{AEs}) generated by white-box attacks on available surrogate models. During the black-box attack stage, DifAttack++ iteratively optimizes the AF based on query feedback from the victim model, while keeping the VF fixed, until a successful AE is obtained. Extensive experimental results demonstrate that DifAttack++ achieves superior ASR and query efficiency compared to state-of-the-art methods, while producing AEs with comparable visual quality. Our code is available at \url{https://github.com/csjunjun/DifAttackPlus.git}. 
\end{abstract}

\begin{IEEEkeywords}
Black box, adversarial attack, query efficient, feature disentanglement, cross domain. 
\end{IEEEkeywords}

\section{Introduction}
\label{sec:intro}
Adversarial Examples ({AE}s) refer to images that deceive Deep Neural Networks ({DNN}s) by incorporating imperceptible perturbations onto clean images. Although AEs pose a security threat to many DNNs~\cite{11267085}, it has also been observed that meticulously crafted AEs are beneficial for evaluating and improving the robustness of DNNs \cite{croce2020reliable,DBLP:journals/tifs/LiLWTZ25,li2024dat,wang2024iterative}, and can even be employed for privacy protection \cite{ric}. The methods for generating AEs can be roughly classified into three categories based on the information of victim DNNs available to the attacker. The first category is the white-box attack \cite{zhang2020walking,11150532}, in which the attacker has access to the architecture and weights of the victim model. The second category is called the black-box attack, where the attacker can only obtain the model's output. The case where the model's output contains both categories and scores is known as a score-based black-box attack~\cite{liu2024boosting,yin2023generalizable}, while the situation where only categories are included is named a decision-based black-box attack \cite{zhao2023sage,chen2023query}. The third category refers to the gray-box attack \cite{9904297}, where the attacker can obtain models' knowledge that falls between white-box attacks and black-box attacks. Given that contemporary potent classification APIs, e.g., \cite{GoogleCloud,Imagga}, generally provide image prediction categories alongside scores, this work primarily focuses on the practical score-based black-box attacks.

We observe that State-Of-The-Art ({SOTA}) score-based black-box attack methods predominantly exhibit two aspects that can be improved. \textbf{On the one hand}, prior techniques that seek AEs by optimizing image features usually operate within the entire feature space, such as AdvFlow \cite{mohaghegh2020advflow} and CGA \cite{feng2022boosting}, without embracing the disentanglement of adversarial capability and visual characteristics in the image feature. Optimizing the entire feature of images could result in substantial pixel perturbations that will encounter truncation by the predefined distortion constraint, thus diminishing the optimization efficacy of the features and eventually increasing query numbers. Another possible chain reaction is the decrease in the Attack Success Rate ({ASR}) caused by the limited query numbers. Although there are a few black-box methods that achieve leading ASR and query efficiency, the visual quality of their generated AEs is often unsatisfactory and quite perceptible. For instance, SquareAttack \cite{andriushchenko2020square} usually produces AEs filled with colored stripes and squares. To improve the unsatisfactory ASR and query efficiency, meanwhile maintaining the visual quality of AEs, our DifAttack++ disentangles an image's latent feature into an Adversarial Feature ({AF}) and a Visual Feature ({VF}), where the former dominates the adversarial capability of an image, while the latter largely determines its visual appearance. Afterward, the AF is iteratively optimized according to the query feedback, while keeping the VF unaltered.

\textbf{On the other hand}, some score-based attack techniques, including P-RGF \cite{cheng2019improving} and GFCS \cite{lord2022attacking}, exploit the gradients from local white-box surrogate models to update AEs during each query. These methods are effective when image categories of the training dataset for surrogate models and the victim model are the same, namely a \textbf{closed-set} black-box attack scenario analogy to the closed-set recognition setting \cite{vaze2021open,wu2023robust}. When it comes to an \textbf{open-set} black-box attack, wherein the training dataset of the victim model is unknown, the effectiveness of these methods declines dramatically, as the possible incongruence between the output categories of surrogate models and the victim model hampers the calculation of gradients from surrogate models. Although there are other approaches, such as SimBA \cite{guo2019simple}, CGA \cite{feng2022boosting} and SquareAttack \cite{andriushchenko2020square}, which are workable in both open-set and closed-set scenarios, they still face challenges in terms of the query efficiency, ASR or severe degradation of image quality. In contrast, our DifAttack++ circumvents the need to compute gradients of surrogate models when updating AEs to be queried. Instead, before embarking on querying the victim model, we leverage AEs generated by performing white-box attack methods on surrogate models to learn how to extract a disentangled AF and VF from the image's latent feature, which enables DifAttack++ to more effectively conduct black-box attacks in open-set scenarios. The efficiency of DifAttack++ in open-set scenarios can be attributed to the fact that the disentangled feature space for optimizing AEs is also responsible for the image reconstruction task, which means that an image can still be reconstructed into its original form with a minimal loss after being mapped into our disentangled feature space. However, the feature space in other methods, such as CGA \cite{feng2022boosting}, is directly used to generate adversarial perturbations that are tightly bound to the data distribution of the training dataset. The latent feature for image reconstructions in DifAttack++ has better generalizability to unknown datasets than the feature space learned from adversarial perturbation distributions of a specific dataset. In addition, DifAttack++ owns the flexibility to selectively harness surrogate models, e.g., in closed-set scenarios, generating transferable AEs as the starting point for the query attack.

More specifically, in DifAttack++, we first train two distinct autoencoders equipped with the designed Hierarchical Decouple-Fusion ({HDF}) module to achieve image reconstructions and to disentangle AFs as well as VFs from images' latent features. In the training stage, our proposed feature interchange operations in Cross-Domain ({CD}), i.e., clean and adversarial image domains, also cooperate with HDF in learning disentangled features. The feature disentanglement is feasible because the adversarial capability is mainly entwined with the decision boundary of classifiers and is relatively independent of the intrinsic image signal, whereas a contrast holds true for the visual appearance. Here, we use two autoencoders with the same architecture but different weights to extract the features of clean images and their AEs separately, since many existing works \cite{stutz2019disentangling,lin2020dual} have pointed out that regular AEs are off the manifold of clean images. It has also been verified that the Intrinsic Dimensionality ({ID}) of AEs is larger than that of clean images in \cite{qing2024detection}, and data with higher ID are more difficult to be modeled by DNNs \cite{pope2020intrinsic}.  Therefore, we reconstruct and disentangle clean images and AEs in CD for more precise feature representations.

After obtaining these well-trained autoencoders, in the attack stage against victim models, we iteratively optimize the AFs of perturbed images using the natural evolution strategy~\cite{wierstra2014natural}, according to the query feedback from the victim model, while keeping the VF unaltered. A successful AE is then reconstructed by fusing the optimized AF with the initial VF. Specifically, the perturbed AFs are extracted from images injected with Gaussian noise. The mean of this sampling distribution is initialized according to the attack scenario: it is set to zero in open-set scenarios, while transferable adversarial perturbations are used in closed-set scenarios. Comprehensive details are provided in the subsequent sections.


Note that an earlier version of DifAttack++, called DifAttack, was published in AAAI 2024~\cite{JunDifAttack2024}. Compared with DifAttack, this work substantially extends the original framework from single-stage, single-domain AF optimization to hierarchical, cross-domain, and transfer-assisted AF optimization. To the best of our knowledge, this is the first work that jointly studies hierarchical AF/VF disentanglement and cross-domain autoencoder learning for query-efficient black-box adversarial attacks. {For clarity, Table~\ref{tab:A1_HDF} provides a systematic comparison between the two versions.} The contributions of this work are summarized as follows:

\begin{itemize}
    \item {We design a Hierarchical Decouple-Fusion (HDF) module to disentangle an image's latent feature into an Adversarial Feature (AF) and a Visual Feature (VF). Different from the one-step DF module in DifAttack, HDF refines preliminary AF/VF features through hierarchical residual-coupling correction before feature fusion.}

    \item {We introduce Cross-Domain (CD) disentanglement with clean-domain and adversarial-domain autoencoders. Beyond using two autoencoders, CD employs attack-oriented reconstruction and disentanglement objectives to encourage VFs to preserve visual appearance and AFs to capture attack-relevant variations across domains.}

    \item {We propose DifAttack++, a query-efficient score-based black-box attack that optimizes AFs according to victim-model feedback while keeping the initial VF invariant. In closed-set scenarios, transferable AEs are disentangled and refined in the learned AF space as a compatible initialization mechanism; in open-set scenarios, the attack starts from Gaussian initialization without using Transferable Initialization (TI).}

    \item {We provide a broader and more challenging evaluation than the AAAI version, including stronger standard classifiers, CLIP-based open-set attacks on Food101, ObjectNet, and VGGFace2 datasets, comprehensive ablation studies, TI analysis, and perceptual-quality evaluation. Experimental results show that DifAttack++ substantially improves ASR and query efficiency over DifAttack and other SOTA score-based black-box attacks while maintaining comparable visual quality.}
\end{itemize}

 The remainder of this paper is organized as follows. Section~\ref{sec:relwork} reviews related works. Section~\ref{sec:method} presents the proposed DifAttack++ framework. Section~\ref{sec:exp} reports experimental results and comparisons with SOTA black-box attacks. It also provides detailed ablation studies to analyze the contributions of HDF, CD, TI, and other key components. Finally, Section~\ref{sec:con} concludes the paper.

\begin{table*}[t!]
\small\setlength{\tabcolsep}{3.0pt}
  \centering
 \caption{{Comparison between the AAAI 2024 DifAttack~\cite{JunDifAttack2024} and current journal extension DifAttack++.}}
 \scalebox{1}{
    \begin{tabular}{p{0.2\linewidth}|p{0.35\linewidth}|p{0.4\linewidth}}
    \hline\hline
    AAAI 2024 DifAttack~\cite{JunDifAttack2024} & TDSC DifAttack++ & Motivation / Intuition \\  \hline
Single-stage Decouple-Fusion (DF) module

&

Two-stage Hierarchical Decouple-Fusion (HDF) module

&

\textbf{Hierarchical feature disentanglement:} Further refines the preliminary Adversarial/Visual Features (AF/VFs) to reduce residual feature coupling.

\\

\hline
A single autoencoder trained for both clean and adversarial image domains
&
Separate clean-domain and adversarial-domain autoencoders jointly constrained by Cross-Domain (CD) reconstruction and disentanglement objectives
&
\textbf{Cross-domain disentanglement:}
Accounts for the manifold and intrinsic-dimensionality differences between clean images and AEs to obtain more precise feature representations across the two domains.

\\

\hline
Random Gaussian noise as the initial perturbation

&

Transferable adversarial perturbations as initialization (TI) in the closed-set setting; Gaussian initialization in the open-set setting

&

\textbf{Attack initialization:}
Leverages transferability to provide a more effective starting point when surrogate models are available, while retaining surrogate-free initialization in the open-set setting.\\

\hline

Evaluation mainly on conventional ImageNet classifiers

&

Expanded evaluation on recent more powerful architectures, CLIP-based models and additional datasets

&

\textbf{Evaluation scope:} Examines the effectiveness and generalization of the attack across more diverse architectures, domains, and practical black-box scenarios.

\\

\hline

Module-level ablation of the DF module

&

Expanded analyses of CD, HDF, and TI, together with parameter-matched comparisons, training-dynamics analysis, semantic-information analysis, perceptual evaluation, practical-cost analysis

&

\textbf{Analysis depth:} Provides a more comprehensive understanding of individual component contributions, architectural design, optimization behavior, semantic properties, perceptual quality, and practical cost.

\\
    \hline\hline
    \end{tabular}%
    }
  \label{tab:A1_HDF}%
\end{table*}%

\section{Related Works}\label{sec:relwork}

\subsection{Score-based Black-box Attack Methods}
\label{sec:relwork:rot}
Current score-based black-box attacks can be broadly categorized into two families according to whether surrogate models are used during the query-based attack stage: surrogate-free and surrogate-assisted attacks.

Surrogate-free attacks \cite{al2020sign, ilyas2018black, li2019nattack, guo2019simple, feng2022boosting, JunDifAttack2024} do not exploit surrogate gradients during querying, and instead rely purely on search strategies guided by query feedback. While these methods exhibit relatively lower ASR and query efficiency, they demonstrate superior generalization in scenarios where surrogate and victim models differ in training data, particularly in open-set settings with disjoint class labels. Among them, a subset of methods, including NES \cite{ilyas2018black}, $\mathcal{N}$Attack \cite{li2019nattack}, and our predecessor DifAttack \cite{JunDifAttack2024}, adopt the natural evolution strategy \cite{wierstra2014natural} for gradient estimation. The primary distinction among these approaches lies in the optimization space, namely the pixel space (NES), the $\tanh$-transformed space ($\mathcal{N}$Attack), or a disentangled adversarial feature space (DifAttack). By operating in a low-dimensional feature space that is highly responsive to adversarial perturbations and optimization, DifAttack achieves superior ASR and query efficiency. 
Although CGA \cite{feng2022boosting} also learns a latent distribution in advance, its Glow \cite{lu2020structured} architecture necessitates downsampling high-resolution inputs (e.g., reducing ImageNet~\cite{deng2009imagenet} images from $224\times224$ to $64\times 64$). This loss of resolution inevitably compromises its attack performance to some extent.

In contrast, surrogate-assisted attacks \cite{cheng2019improving, guo2019subspace, cai2022blackbox, lord2022attacking, yin2023generalizable} explicitly leverage surrogate model gradients during the query phase to guide the search. For instance, methods such as P-RGF \cite{cheng2019improving} and Subspace \cite{guo2019subspace} bias their search directions toward surrogate gradients, while others, including BASES \cite{cai2022blackbox} and MCG \cite{yin2023generalizable}, adaptively select or finetune surrogate information based on query feedback. The SOTA MCGSquare \cite{yin2023generalizable} further integrates the meta-learning framework of MCG with the Square~\cite{andriushchenko2020square} search strategy. Despite their effectiveness in certain settings, surrogate-assisted attacks suffer from two limitations. 
First, they rely on category consistency between surrogate and victim models, leading to degradations in open-set scenarios where data distributions mismatch and surrogate gradients become unreliable. Second, achieving an optimal trade-off between attack capability, query efficiency, and visual quality remains a challenge for these methods.

\subsection{Disentangled Representation (DR)}

We now give a brief introduction of DR, which aims to separate the distinct, informative generative factors of data \cite{bengio2013representation}. Modern approaches often employ autoencoders \cite{DRIT_plus}, GANs \cite{meo2023alpha}, or other generative models \cite{ma2020decoupling} to learn this separation, frequently optimizing metrics like mutual information \cite{burgess2018understanding} or total correlation \cite{kim2018disentangling}. In the security domain, DR is most commonly harnessed for AE detection or defense \cite{mustafa2020deeply,wang2021defending,yang2021adversarial,qing2024detection}. To our knowledge, the few attack methods involving DR are exclusively white-box approaches. For instance, RNF \cite{kim2021distilling} optimizes perturbations against classifier-specific ``non-robust features", while SSAE \cite{lu2021discriminator} trains an autoencoder to directly generate AEs for a specific victim. These methods require full knowledge of the victim model's parameters to learn their respective DRs. Unlike white-box methods \cite{kim2021distilling}, our approach operates in black-box settings where the victim models’ parameters and architectures are completely unknown. It learns a generalizable DR from surrogate models that reconstructs the image signal itself, rather than classifier-specific features.

\subsection{On the Distribution of Adversarial Examples}
\label{sec:relwork:dist}

A significant body of research indicates that AEs are not mere outliers, but instead occupy a data distribution that is distinct from that of natural, clean images. Seminal works \cite{stutz2019disentangling, lin2020dual} have demonstrated that AEs often lie ``off-manifold'', residing in low-probability regions of the natural image distribution. This discrepancy is further supported by studies on ID; for example, \cite{qing2024detection} and \cite{pope2020intrinsic} show that AEs exhibit higher ID and are more complex to model than their clean counterparts.

Despite these theoretical insights, most existing generative attack methods (e.g., CGA \cite{feng2022boosting} and our predecessor DifAttack \cite{JunDifAttack2024}) implicitly assume a shared latent manifold, employing a single generative model to represent both clean and adversarial distributions. We argue that modeling these fundamentally different distributions with a shared backbone may be sub-optimal. In contrast, DifAttack++ explicitly introduces a cross-domain architecture that respects the distinct manifold properties of clean and adversarial data.

\begin{figure*}[t!]
\centering
\includegraphics[width=1\textwidth]{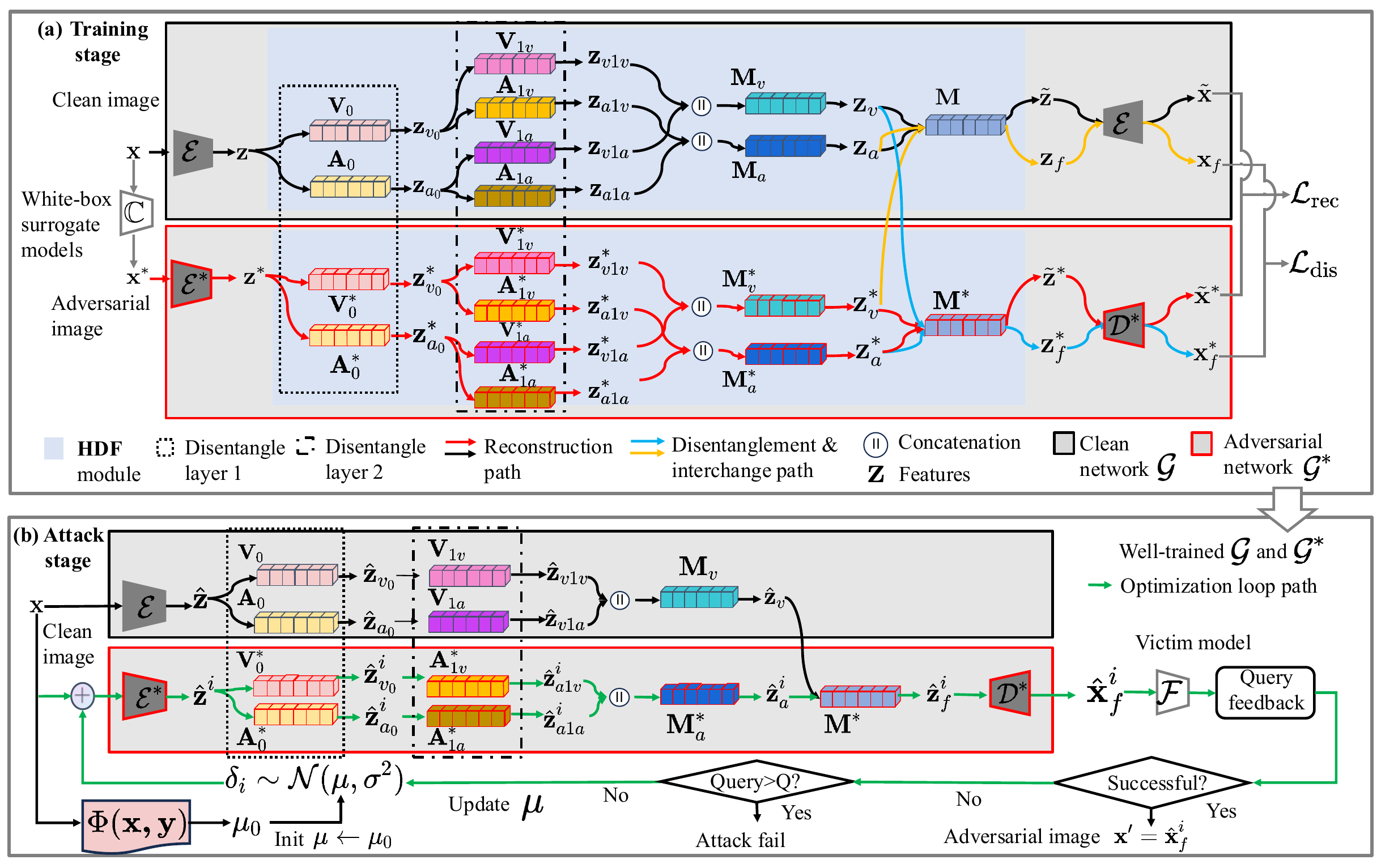} 
\caption{
(a) The training procedure of the cross-domain autoencoders $\mathcal{G}$ and $\mathcal{G}^*$, equipped with the proposed HDF module to disentangle AFs ($\mathbf{z}_a$) and VFs ($\mathbf{z}_v$), optimized by minimizing the combined image reconstruction loss $\mathcal{L}_{rec}$ and feature disentanglement loss $\mathcal{L}_{dis}$.
(b) The proposed score-based black-box attack method, \emph{DifAttack++}, which leverages the pre-trained $\mathcal{G}$ and $\mathcal{G}^*$ to extract VFs ($\hat{\mathbf{z}}_v$) and AFs ($\hat{\mathbf{z}}_a^i$), keeps the VF fixed, and iteratively optimizes the AF until a successful AE ($\mathbf{x}^\prime$) is found or the query budget ($Q$) is exhausted.
}
\label{fig1}
\end{figure*}

\section{Proposed DifAttack++}
\label{sec:method}

\subsection{Overview}
\label{sec:method:overview}
DifAttack++ aims to exploit disentangled VFs and AFs to decode AEs in score-based black-box settings, guided by query feedback from the victim model. As illustrated in Fig.~\ref{fig1}, DifAttack++ consists of two main phases.

In the training stage, we train two U-Net-like \cite{ronneberger2015u} autoencoders, denoted as $\mathcal{G}$ and $\mathcal{G}^*$, which share the same architecture but differ in their weights. Both autoencoders are capable of feature disentanglement and image reconstruction. Specifically, $\mathcal{G}$ is trained on clean images $\mathbf{x}$, while $\mathcal{G}^*$ is trained on their corresponding adversarial examples $\mathbf{x}^*$, where $\mathbf{x}^*$ is generated by applying untargeted white-box attacks on surrogate models. Taking a clean image $\mathbf{x}$ as an example, $\mathcal{G}$ first disentangles its latent representation $\mathbf{z}$ into an AF $\mathbf{z}_a$ and a VF $\mathbf{z}_v$ via the proposed HDF module, and then reconstructs an image $\tilde{\mathbf{x}}$ that is similar to $\mathbf{x}$ by decoding the fused features $\mathbf{z}_a$ and $\mathbf{z}_v$. Similarly, $\mathcal{G}^*$ performs the same operations for adversarial inputs.

In the attack stage, we leverage the pre-trained autoencoders $\mathcal{G}$ and $\mathcal{G}^*$ to generate an adversarial example $\mathbf{x}^\prime$ for a given clean image $\mathbf{x}$ against a black-box victim model $\mathcal{F}$. Specifically, $\mathbf{x}^\prime$ is obtained through an iterative optimization process. At each iteration, we sample a batch of perturbed images to extract their corresponding AFs $\hat{\mathbf{z}}_a^i$, while keeping the VF $\hat{\mathbf{z}}_v$—which is used for subsequent image reconstruction together with $\hat{\mathbf{z}}_a^i$—fixed and identical to that of the clean image. This procedure produces a batch of reconstructed images $\hat{\mathbf{x}}_f^i$ that are queried against $\mathcal{F}$. The process continues until either the query budget $Q$ is exhausted or one of the reconstructed images $\hat{\mathbf{x}}_f^i$ successfully deceives $\mathcal{F}$, in which case we obtain $\mathbf{x}^\prime = \hat{\mathbf{x}}_f^i$.

\subsection{Training Stage: Train Autoencoders \texorpdfstring{$\mathcal{G}$ and $\mathcal{G}^*$}{G and G*}}
\label{sec:trainstage}

Let $\mathbf{x}$ denote a clean image from the training dataset $\mathbb{T}$, and let $\mathbb{C}$ represent the set of pre-trained white-box surrogate models trained on $\mathbb{T}$. Note that the image classes and data distribution of $\mathbb{T}$ may differ from those of the test dataset. To generate an AE $\mathbf{x}^*$ corresponding to $\mathbf{x}$, we adopt the white-box attack method PGD~\cite{madry2018towards} by attacking a surrogate model $\mathcal{C}_j$, which is randomly selected from $\mathbb{C}$. Clearly, PGD can be replaced by other white-box attack methods, or by a combination of such methods, without significantly affecting the final attack performance of our method (see the supplementary material of DifAttack~\cite{JunDifAttack2024} for more details). Therefore, in this work, we adopt PGD as a representative choice. Unlike our previous work DifAttack, which employs a single autoencoder shared by both clean and adversarial images, we instead assign two distinct autoencoders, $\mathcal{G}$ and $\mathcal{G}^*$, to the clean and adversarial domains, respectively. Notably, regular AEs usually lie off the manifold of normal images~\cite{lin2020dual}. In this regard, adopting separate autoencoders is more appropriate and can lead to improved training efficiency and enhanced black-box attack performance, as will be demonstrated experimentally. In what follows, we provide a detailed description of the disentanglement and reconstruction tasks performed by the two autoencoders.

\subsubsection{Disentanglement} 
To identify AFs to which an image’s adversarial capability is sensitive, while ensuring that modifications to these features minimally affect visual appearance, we disentangle AFs and VFs from the image’s latent representation. To this end, we exchange AFs and VFs between a clean image and its corresponding AE. The two images reconstructed from the exchanged features are then expected to exhibit adversarial capabilities and visual appearances that correspond to the images from which their respective features are derived.

\noindent{\textbf{HDF Module.}} 
To implement feature disentanglement, as illustrated in Fig.~\ref{fig1}(a), we design a {Hierarchical Decouple-Fusion} (HDF) module. The HDF module takes an image's latent representation as input, disentangles it into VFs and AFs, and subsequently reconstructs the latent representation by fusing the disentangled features. The HDF module comprises three types of convolutional layers: visual layers ($\mathbf{V}_0$, $\mathbf{V}_{1v}$, and $\mathbf{V}_{1a}$), adversarial layers ($\mathbf{A}_0$, $\mathbf{A}_{1v}$, and $\mathbf{A}_{1a}$), and fusion layers ($\mathbf{M}_a$, $\mathbf{M}_v$, and $\mathbf{M}$).

All these layers are implemented using $1\times1$ convolutions. 
{To clarify the role of these lightweight projections, we first consider the decoupling projections in HDF. For any decoupling layer 
\(\mathbf{P}\in\{\mathbf{V}_0,\mathbf{A}_0,\mathbf{V}_{1v},\mathbf{V}_{1a},\mathbf{A}_{1v},\mathbf{A}_{1a}\}\) 
and an input feature \(\mathbf{u}\in\mathbb{R}^{C\times H\times W}\), the $1\times1$ convolution maps each channel vector independently as}
\begin{equation}\label{eq:conv1x1}
{
\mathbf{P}(\mathbf{u})_{h,w}
=
\mathbf{W}_{\mathbf{P}}\mathbf{u}_{h,w}
+
\mathbf{b}_{\mathbf{P}},
}
\end{equation}
{where \(\mathbf{u}_{h,w}\in\mathbb{R}^{C}\), 
\(\mathbf{W}_{\mathbf{P}}\in\mathbb{R}^{C'\times C}\), and 
\(\mathbf{b}_{\mathbf{P}}\in\mathbb{R}^{C'}\). 
Here, \(C'\) denotes the number of output channels. 
Thus, the decoupling layers learn feature-specific channel projections without introducing additional spatial filtering, while the fusion layers \(\mathbf{M}_v\), \(\mathbf{M}_a\), and \(\mathbf{M}\) subsequently fuse the calibrated components.}

Taking a clean image $\mathbf{x}$ as an example, its latent representation is $\mathbf{z} = \mathcal{E}(\mathbf{x})$, where $\mathcal{E}$ denotes a convolutional neural network (CNN)-based encoder. The representation $\mathbf{z}$ is first fed into the initial disentanglement layers $\mathbf{V}_0$ and $\mathbf{A}_0$ to obtain preliminary visual features $\mathbf{z}_{v0}$ and adversarial features $\mathbf{z}_{a0}$:
\begin{equation}\label{eq:layer1}
\begin{aligned}
\mathbf{z}_{v0}=\mathbf{V}_{0}(\mathbf{z}),\quad
\mathbf{z}_{a0}=\mathbf{A}_{0}(\mathbf{z}).
\end{aligned}
\end{equation}
{This first stage provides a coarse AF/VF decomposition. 
However, because the separation is performed by a single pair of projections, the preliminary features may still contain residual coupling; for example, $\mathbf{z}_{v0}$ may contain adversarially sensitive information, while $\mathbf{z}_{a0}$ may retain visual-preserving components.}

{To refine this coarse separation, HDF introduces a second hierarchical disentanglement stage between the preliminary projections and the final fusion.} 
Specifically, four additional disentanglement layers---$\mathbf{V}_{1v}$, $\mathbf{V}_{1a}$, $\mathbf{A}_{1v}$, and $\mathbf{A}_{1a}$---are appended to extract fine-grained VFs and AFs from the preliminary representations:
\begin{equation}\label{eq:layer2}
\begin{aligned}
\mathbf{z}_{v1v}=\mathbf{V}_{1v}(\mathbf{z}_{v0}),\quad
\mathbf{z}_{v1a}=\mathbf{V}_{1a}(\mathbf{z}_{a0}),\\
\mathbf{z}_{a1v}=\mathbf{A}_{1v}(\mathbf{z}_{v0}),\quad
\mathbf{z}_{a1a}=\mathbf{A}_{1a}(\mathbf{z}_{a0}).
\end{aligned}
\end{equation}
{These four paths explicitly model same-feature preservation and cross-feature correction. 
The paths $\mathbf{z}_{v0}\!\rightarrow\!\mathbf{z}_{v1v}$ and $\mathbf{z}_{a0}\!\rightarrow\!\mathbf{z}_{a1a}$ preserve feature-consistent information. 
In contrast, $\mathbf{z}_{a0}\!\rightarrow\!\mathbf{z}_{v1a}$ extracts residual visual-relevant components from the preliminary AF, while $\mathbf{z}_{v0}\!\rightarrow\!\mathbf{z}_{a1v}$ extracts residual adversarial-relevant components from the preliminary VF.}

The hierarchically calibrated VFs $\mathbf{z}_{v1v}$ and $\mathbf{z}_{v1a}$ are then combined through a visual fusion layer $\mathbf{M}_v$ to produce the final disentangled VF $\mathbf{z}_v$ of $\mathbf{x}$. Similarly, an adversarial fusion layer $\mathbf{M}_a$ integrates $\mathbf{z}_{a1v}$ and $\mathbf{z}_{a1a}$ to yield the final AF $\mathbf{z}_a$:
\begin{equation}\label{eq:fusionlayer}
\begin{aligned}
\mathbf{z}_v &= \mathbf{M}_v\big(\mathbf{z}_{v1v} \,||\, \mathbf{z}_{v1a}\big), \\
\mathbf{z}_a &= \mathbf{M}_a\big(\mathbf{z}_{a1v} \,||\, \mathbf{z}_{a1a}\big),
\end{aligned}
\end{equation}
where $||$ denotes channel-wise concatenation. 
{Therefore, the second stage can be interpreted as residual coupling correction before fusion. It does not simply stack another $1\times1$ convolution on every branch; instead, it inserts a structured intermediate feature decomposition between preliminary disentanglement and final fusion. This decomposition separates same-feature preservation from cross-feature reassignment, and the calibrated components are then recombined through the branch-specific fusion layers $\mathbf{M}_v$ and $\mathbf{M}_a$.}

Subsequently, $\mathbf{z}_v$ and $\mathbf{z}_a$ are fused through the fusion layer $\mathbf{M}$ to reconstruct the latent representation $\tilde{\mathbf{z}}$. Overall, the feature processing of the HDF module can be summarized as
\begin{equation}\label{eq:finalfusion}
\begin{aligned}
\mathcal{HDF}(\mathbf{z}) = \mathbf{M}\big(\mathbf{z}_v \,||\, \mathbf{z}_a\big) = \tilde{\mathbf{z}}.
\end{aligned}
\end{equation}
For clarity, we denote the HDF modules in $\mathcal{G}$ and $\mathcal{G}^*$ as $\mathcal{HDF}(\cdot)$ and $\mathcal{HDF}^*(\cdot)$, respectively. The corresponding fusion process in $\mathcal{G}^*$ is defined analogously for the latent representation $\mathbf{z}^*$ of an AE $\mathbf{x}^*$:
\begin{equation}\label{eq:defu1}
\begin{aligned}
\mathcal{HDF}^*(\mathbf{z}^*) = \mathbf{M}^*\big(\mathbf{z}_v^* \,||\, \mathbf{z}_a^*\big) = \tilde{\mathbf{z}}^*,
\end{aligned}
\end{equation}
where $\mathbf{M}^*$ denotes the fusion layer in $\mathcal{G}^*$.

\noindent{\textbf{Feature Interchange in CD.}} With the HDF module in place, we now describe how feature interchange is performed in the cross-domain (CD) to facilitate effective disentanglement. Specifically, as illustrated by the yellow path in Fig.~\ref{fig1}(a), the AFs of a clean image $\mathbf{z}_a$ are concatenated with the VFs of an adversarial image $\mathbf{z}_v^*$, and the resulting features are fed into the fusion layer $\mathbf{M}$ to obtain a fused representation $\mathbf{z}_f$. Symmetrically, as shown by the blue path in Fig.~\ref{fig1}(a), another fused representation $\mathbf{z}_f^*$ is obtained by combining the VFs of the clean image $\mathbf{z}_v$ with the AFs of the adversarial image $\mathbf{z}_a^*$:
\begin{equation}\label{eq:mixf}
\begin{aligned}
&\mathbf{z}_f=\mathbf{M}(\mathbf{z}_v^*||\mathbf{z}_a), \quad 
&\mathbf{z}_f^*=\mathbf{M}^*(\mathbf{z}_v||\mathbf{z}_a^*).
\end{aligned}
\end{equation} 
Given the fused features, the corresponding reconstructed images $\mathbf{x}_f$ and $\mathbf{x}_f^*$ are then obtained as
\begin{equation}\label{eq:reconmixx}
\begin{aligned}
&\mathbf{x}_f=\mathcal{D}(\mathbf{z}_f),\quad
&\mathbf{x}_f^*=\mathcal{D}^*(\mathbf{z}_f^*),
\end{aligned}
\end{equation} 
where $\mathcal{D}$ and $\mathcal{D}^*$ denote the CNN-based decoders in $\mathcal{G}$ and $\mathcal{G}^*$, respectively.

 \noindent{\textbf{Disentanglement Loss.}} 
Subsequently, to enforce that AFs primarily govern the adversarial capability of an image, while VFs mainly determine its perceptual appearance, we impose complementary constraints on the reconstructed images. Specifically, the fused image $\mathbf{x}_f$ is expected to be visually similar to the adversarial example $\mathbf{x}^*$ while remaining non-adversarial. In contrast, $\mathbf{x}_f^*$ should preserve the visual appearance of the clean image $\mathbf{x}$, yet remain adversarial against all surrogate models to enhance the generalization of AFs across different classifiers. Accordingly, we define the disentanglement loss ${L}_{dis}$ as
\begin{equation}\label{eq:disenloss}
\begin{gathered}
{L}_{dis} = ||\mathbf{x}^*-\mathbf{x}_f||_2+\frac{1}{N}\sum\nolimits_{j=1}^{N}{L}_{adv}(\mathbf{x}_f,y,1,\mathcal{C}_j,k)\\
+||\mathbf{x}-\mathbf{x}_f^*||_2+\frac{1}{N}\sum\nolimits_{j=1}^{N}{L}_{adv}(\mathbf{x}_f^*,y,0,\mathcal{C}_j,k),
\end{gathered}
\end{equation}
where $N$ denotes the cardinality of the surrogate model set $\mathbb{C}$. The $\ell_2$ norm measures the Euclidean distance between two images, and ${L}_{adv}$ quantifies adversarial effectiveness. Following \cite{carlini2017towards}, we employ the margin loss for optimization. Unlike cross-entropy, this formulation allows us to explicitly control the attack confidence $k$. This loss is formulated as:
\begin{equation}\label{eq:advloss}
\begin{gathered}
{L}_{adv}(\mathbf{x}, y, v, \mathcal{C}_j, k) = \\
\max\Big\{ \mathbb{I}(v)\!\cdot\!\big( 
\mathcal{C}_j(\mathbf{x}, y) - \max_{h \neq y} \mathcal{C}_j(\mathbf{x}, h) 
\big),\; -k \Big\}.
\end{gathered}
\end{equation}
Here, $v \in \{0,1\}$ specifies the attack type. Specifically, $\mathbb{I}(0)=1$ corresponds to an untargeted attack with ground-truth label $y$, whereas $\mathbb{I}(1)=-1$ indicates a targeted attack with target class $y$. Moreover, $\mathcal{C}_j(\mathbf{x}, h)$ denotes the output score of classifier $\mathcal{C}_j$ for class $h$.

\subsubsection{Reconstruction} 
To enforce cycle consistency \cite{zhu2017unpaired} in the context of feature disentanglement, as illustrated by the reconstruction paths in Fig.~\ref{fig1}(a), the autoencoder is required to reconstruct an input image well even after its latent representation has been processed by the HDF module. Accordingly, the outputs of the HDF modules, denoted as $\tilde{\mathbf{z}}$ and $\tilde{\mathbf{z}}^*$, are expected to reconstruct images $\tilde{\mathbf{x}} \approx \mathbf{x}$ and $\tilde{\mathbf{x}}^* \approx \mathbf{x}^*$, respectively, via
\begin{equation}\label{eq:reconx}
\begin{aligned}
&\tilde{\mathbf{x}}=\mathcal{D}(\mathcal{HDF}(\mathbf{z}) ),\quad
&\tilde{\mathbf{x}}^*=\mathcal{D}^*(\mathcal{HDF}^*(\mathbf{z}^*)).
\end{aligned}
\end{equation} 
Based on these reconstructions, we define the reconstruction loss ${L}_{rec}$ for networks $\mathcal{G}$ and $\mathcal{G}^*$ as
\begin{equation}\label{eq:reconloss}
\begin{aligned}
{L}_{rec} = \|\mathbf{x} - \tilde{\mathbf{x}}\|_2 
+ \|\mathbf{x}^* - \tilde{\mathbf{x}}^*\|_2 .
\end{aligned}
\end{equation}
Finally, the overall objective optimized during the training of $\mathcal{G}$ and $\mathcal{G}^*$ is given by
\begin{equation}\label{eq:totalloss}
\begin{aligned}
{L}_{all} = \lambda \cdot {L}_{rec} + {L}_{dis},
\end{aligned}
\end{equation}
where the hyperparameter $\lambda$ controls the trade-off between the reconstruction and disentanglement losses.

\subsection{Attack Stage: Generating the AE \texorpdfstring{$\mathbf{x}^\prime$}{x'}}
\label{sec:attackstage}

\noindent{\textbf{VF}.} After obtaining the well-trained models $\mathcal{G}$ and $\mathcal{G}^*$, we leverage them in the attack stage to generate AEs under a black-box setting. As illustrated in Fig.~\ref{fig1}(b), given a clean image $\mathbf{x}$, our goal is to attack the victim model $\mathcal{F}$ while preserving the visual fidelity of the generated adversarial image. To this end, we keep the VF used for decoding the perturbed image $\hat{\mathbf{x}}_f^i$ identical to that of the original image $\mathbf{x}$. Specifically, the VF $\hat{\mathbf{z}}_v$ is extracted as
\begin{equation}\begin{gathered}\label{eq:open1}
\hat{\mathbf{z}}=\mathcal{E}({\mathbf{x}}),\\
\hat{\mathbf{z}}_{v0}=\mathbf{V}_0(\hat{\mathbf{z}}),\quad
\hat{\mathbf{z}}_{a0}=\mathbf{A}_0(\hat{\mathbf{z}}),\\
\hat{\mathbf{z}}_{v1v}=\mathbf{V}_{1v}(\hat{\mathbf{z}}_{v0}),\quad
\hat{\mathbf{z}}_{v1a}=\mathbf{V}_{1a}(\hat{\mathbf{z}}_{a0}),\\
\hat{\mathbf{z}}_v=\mathbf{M}_v(\hat{\mathbf{z}}_{v1v}||\hat{\mathbf{z}}_{v1a}).
\end{gathered}
\end{equation}
This design encourages $\hat{\mathbf{x}}_f^i$ to remain as close as possible to the clean image in terms of perceptual appearance.

\noindent{\textbf{AF}.} With the VF fixed, the next objective is to identify an AF $\hat{\mathbf{z}}_a^i$ such that, when combined with $\hat{\mathbf{z}}_v$, it can successfully deceive the victim model $\mathcal{F}$. For instance, in the untargeted attack setting, we aim to generate an adversarial image $\hat{\mathbf{x}}_f^i$ by solving
\begin{equation}
\begin{gathered}\label{eq:attackobj}
\hat{\mathbf{x}}_f^i=\mathcal{D}^*(\mathbf{M}^*(\hat{\mathbf{z}}_v||\hat{\mathbf{z}}_a^i)),\\
s.t.\ \arg\max_{h}\mathcal{F}(\hat{\mathbf{x}}_f^i,h)\neq y,\quad
\|\hat{\mathbf{x}}_f^i-\mathbf{x}\|_p\leq\epsilon,
\end{gathered}
\end{equation}
where $y$ denotes the ground-truth label of $\mathbf{x}$. For targeted attacks, the first constraint is modified to $\arg\max_{h}\mathcal{F}(\hat{\mathbf{x}}_f^i,h)=y$, where $y$ now represents the target class.

To obtain a suitable AF ${\mathbf{z}}_a^i$, we search for a perturbation $\bm\delta_i$ in the pixel space such that the resulting perturbed image ${\mathbf{x}}+\bm\delta_i$ induces adversarially effective AFs. This design choice is motivated by the fact that the pixel space allows more direct and precise control over the perturbation magnitude than the feature space. To alleviate the curse of dimensionality and exploit local gradient similarity~\cite{ilyas2019prior}, the optimization is conducted in a downsampled pixel space $\mathbb{R}^{C \times (H/s) \times (W/s)}$ with scaling factor $s$. Formally, $\hat{\mathbf{z}}_a^i$ is extracted as
\begin{equation}\begin{gathered}\label{eq:adv1}
\hat{\mathbf{z}}^i=\mathcal{E}^*({\mathbf{x}}+\operatorname{up}(\bm\delta_i,s)),\\
\hat{\mathbf{z}}_{v0}^i=\mathbf{V}_0^*(\hat{\mathbf{z}}^i),\quad
\hat{\mathbf{z}}_{a0}^i=\mathbf{A}_0^*(\hat{\mathbf{z}}^i),\\
\hat{\mathbf{z}}_{a1v}^i=\mathbf{A}_{1v}^*(\hat{\mathbf{z}}_{v0}^i),\quad
\hat{\mathbf{z}}_{a1a}^i=\mathbf{A}_{1a}^*(\hat{\mathbf{z}}_{a0}^i),\\
\hat{\mathbf{z}}_a^i=\mathbf{M}_a^*(\hat{\mathbf{z}}_{a1v}^i||\hat{\mathbf{z}}_{a1a}^i),
\end{gathered}
\end{equation}
where $\operatorname{up}(\cdot)$ denotes the upsampling operation that projects $\bm\delta_i$ back to the original input resolution.
The candidate adversarial image is then reconstructed as
\begin{equation}\begin{aligned}\label{eq:adv2}
\hat{\mathbf{x}}_f^i=\Pi_{\epsilon,\mathbf{x}}\!\left(
\mathcal{D}^*\big(\mathbf{M}^*(\hat{\mathbf{z}}_v||\hat{\mathbf{z}}_a^i)\big)
\right),
\end{aligned}
\end{equation}
where $\hat{\mathbf{z}}_v$ is fixed as defined in Eq.~(\ref{eq:open1}), and $\hat{\mathbf{z}}_a^i$ is iteratively optimized through Eq.~(\ref{eq:adv1}) by updating $\bm\delta_i$. The operator $\Pi_{\epsilon,\mathbf{x}}$ projects the image onto the $\ell_p$-ball of radius $\epsilon$ centered at $\mathbf{x}$, ensuring that the perturbation satisfies the prescribed constraint. For simplicity, we denote the right-hand side of Eq.~(\ref{eq:adv2}) as $\mathbf{T}({\mathbf{x}}+\operatorname{up}(\bm\delta_i))$.

By combining Eq.~(\ref{eq:adv2}) with a change-of-variables formulation, we unify the attack objective in Eq.~(\ref{eq:attackobj}) as
\begin{equation}\begin{gathered}\label{eq:algo3}
\min_{\bm\mu}~\mathbb{E}_{\mathcal{N}(\bm\delta_i|\bm\mu,\sigma^2)}
L_{adv}(\mathbf{T}({\mathbf{x}}+\operatorname{up}(\bm\delta_i)),y,v,\mathcal{F},k),\\
s.t.\ \|\mathbf{T}({\mathbf{x}}+\operatorname{up}(\bm\delta_i))-\mathbf{x}\|\leq \epsilon,
\end{gathered}
\end{equation}
where $\hat{\mathbf{x}}_f^i=\mathbf{T}(\hat{\mathbf{x}}+\operatorname{up}(\bm\delta_i))$ satisfies the untargeted or targeted attack condition accordingly. We sample $\bm\delta_i$ from a Gaussian distribution $\mathcal{N}(\bm\mu,\sigma^2)$, where the mean $\bm\mu$ is optimized and the variance $\sigma$ is fixed based on a grid search. To optimize $\bm\mu$, we adopt the natural evolution strategy \cite{wierstra2014natural} to estimate the gradient of the expected adversarial loss with respect to $\bm\mu$. Using stochastic gradient descent with learning rate $\eta$ and batch size $\tau$, the update rule is
\begin{equation}\begin{aligned}\label{eq:algo4}
\bm\mu \leftarrow \bm\mu-\frac{\eta}{\tau\sigma}\sum_{i=1}^{\tau}
\bm\gamma_i\,L_{adv}(\mathbf{T}({\mathbf{x}}+\operatorname{up}(\bm\mu+\sigma\bm\gamma_i)),y,v,\mathcal{F},k),
\end{aligned}
\end{equation}
where $\bm\gamma_i\sim\mathcal{N}(\bm0,I)$. Once a feasible solution for $\bm\mu$ is obtained, a successful AE $\mathbf{x}^\prime$ can be sampled as
\begin{equation}\begin{aligned}\label{eq:algo5}
\mathbf{x}^\prime=\hat{\mathbf{x}}_f^i=\mathbf{T}({\mathbf{x}}+\operatorname{up}(\bm\mu+\sigma\bm\gamma_i)).
\end{aligned}
\end{equation}

However, the convergence and effectiveness of this optimization depend on the initialization of $\bm\mu$. Poor initialization may lead to slow convergence or make it difficult to find a valid perturbation under the budget constraint. We therefore adopt different initialization strategies for open-set and closed-set scenarios, denoted as \textbf{${\text{Dif++}_{\text{open}}}$} and \textbf{${\text{Dif++}_{\text{close}}}$}, respectively.

In the open-set setting, the surrogate and victim models are trained on different label spaces or even different image domains. As a result, transferable AEs generated from surrogate models may not provide reliable warm starts for the victim model. Therefore, {Dif++$_{\text{open}}$ does not use Transferable Initialization (TI) during the attack stage}; instead, it initializes the perturbation from zero-mean Gaussian noise. In the closed-set setting, the surrogate and victim models share the same label space, so adversarial transferability is more reliable. We thus exploit transferable AEs as an adversarially biased initialization. The two cases are unified as
\begin{equation}\label{eq:unimuinit}
\bm\mu_0 =
\begin{cases}
\mathbf{0}, & \text{open-set}, \\
\operatorname{down}(\mathcal{A}(\mathbf{x},y,\mathbb{C},\epsilon),s), & \text{closed-set},
\end{cases}
\end{equation}
which we denote as $\Phi(\mathbf{x}, y)$.

Here, $\operatorname{down}(\cdot)$ denotes a downsampling operator with scale $s$, and $\mathcal{A}$ represents a transferable adversarial attack algorithm that, given a clean image $\mathbf{x}$, its label $y$, a set of surrogate models $\mathbb{C}$, and a perturbation budget $\epsilon$, produces a perturbed image within the allowed budget. We instantiate $\mathcal{A}$ using PGN~\cite{ge2023boosting} for untargeted attacks and FTM~\cite{liang2025improving} for targeted attacks due to their strong transferability. $\mathcal{A}$ is flexible and can be replaced with stronger methods in the future. {Note that $\mathcal{A}$ is used only in the closed-set setting. In the open-set setting, no attack-stage surrogate model is required, although the autoencoders still need to be trained beforehand using surrogate models to learn the AF/VF disentangled space.}

\noindent{\textbf{Zero-Shot AE Check}}. Prior to the iterative optimization of $\bm\mu$, {we conduct a preliminary verification to check whether the initial candidate, reconstructed from TI in the closed-set setting or Gaussian initialization in the open-set setting, already becomes adversarial through the learned AF/VF space.} Specifically, we generate a candidate AE via Eq.~(\ref{eq:adv2}) and query the victim model. If this initial candidate successfully misleads the model, the attack terminates immediately, thereby significantly reducing query costs. Otherwise, we proceed to the iterative optimization of $\bm\mu$ as formulated in Eqs.~(\ref{eq:algo3})-(\ref{eq:algo4}). This mechanism effectively enables a potential ``zero-shot'' attack. The complete workflow is summarized in Algorithm~\ref{alg:algorithm}, with this verification step detailed in Lines~3-7. Additionally, following~\cite{li2019nattack}, we normalize the adversarial loss $L_{adv}$ (Line~20) to stabilize convergence.

{\color{black}
\begin{algorithm}[tb]
\color{black}
\small
\caption{DifAttack++}
\label{alg:algorithm}
\textbf{Input:} Target classifier $\mathcal{F}$; clean image $\mathbf{x}$; query budget $Q$; perturbation budget $\epsilon$; ground-truth/target label $y$ (for untargeted/targeted attacks); learning rate $\eta$; variance $\sigma$; parameters $v$ and $k$ in Eq.~(\ref{eq:algo3}); sample size $\tau$; downsampling/upsampling factor $s$.\\
\textbf{Output:} Adversarial image $\mathbf{x}^\prime$.
\begin{algorithmic}[1]
\STATE Initialize query counter $q \gets 1$
\STATE Initialize $\bm\mu_0 \gets \mathbf{0}$ for open-set scenarios (\textbf{$\text{Dif++}_{\text{open}}$}), \\ or set $\bm\mu_0$ via Eq.~(\ref{eq:unimuinit}) for closed-set scenarios (\textbf{$\text{Dif++}_{\text{close}}$)}
\STATE Sample initial mean $\bm\mu \sim \mathcal{N}(\bm\mu_0, I)$
\STATE Generate an initial adversarial candidate $\hat{\mathbf{x}}_f^0$ using Eq.~(\ref{eq:adv2})
\IF{${L}_{adv}(\hat{\mathbf{x}}_f^0, y, v, \mathcal{F}, k) = -k$}
    \STATE \textbf{return} $\mathbf{x}^\prime \gets \hat{\mathbf{x}}_f^0$
\ENDIF
\WHILE{$q + \tau \leq Q$}
    \STATE Sample $\gamma_1, \ldots, \gamma_\tau \sim \mathcal{N}(\bm0, I)$
    \STATE Set $\bm\delta_i \gets \bm\mu + \sigma \gamma_i$, $\forall i \in \{1,\ldots,\tau\}$
    \STATE Compute adversarial candidates $\hat{\mathbf{x}}_f^i$ using Eqs.~(\ref{eq:open1}), (\ref{eq:adv1}), and (\ref{eq:adv2})
    \STATE Evaluate $l_i \gets {L}_{adv}(\hat{\mathbf{x}}_f^i, y, v, \mathcal{F}, k)$, $\forall i$
    \STATE Update query count $q \gets q + \tau$
    \IF{$\exists\, i \text{ s.t. } l_i = -k$}
        \STATE \textbf{return} $\mathbf{x}^\prime \gets \hat{\mathbf{x}}_f^i$
        \STATE \textbf{break}
    \ELSE
        \STATE Normalize losses $\hat{l}_i \gets \big(l_i - \mathrm{mean}(\{l_i\})\big) / \mathrm{std}(\{l_i\})$, $\forall i$
        \STATE Update mean $\bm\mu \gets \bm\mu - \frac{\eta}{\tau\sigma} \sum_{i=1}^{\tau} \hat{l}_i \gamma_i$
    \ENDIF
\ENDWHILE
\STATE \textbf{return} $\mathbf{x}^\prime$
\end{algorithmic}
\end{algorithm}
}

\begin{table*}[t!]
\small\tabcolsep=0.02cm
  \centering
  \caption{Results of score-based black-box attacks on ImageNet for the ``\textit{simple group}''. The attack success rate (ASR, \%) as well as the average and median numbers of queries (Avg.Q/Med.Q) are reported as evaluation metrics. The best and second-best results are highlighted in \textbf{bold} and \underline{underlined}, respectively.}
  \scalebox{0.75}{
    \begin{tabular}{c|c|ccc|ccc|ccc|ccc|ccc|ccc|ccc|ccc}
    \hline\hline
    
    \multirow{3}[2]{*}{Scenarios} &     Victims & \multicolumn{6}{c|}{GoogleNet}    & \multicolumn{6}{c|}{VGG16}  & \multicolumn{6}{c|}{ResNet18} & \multicolumn{6}{c}{SqueezeNet}  \\
    
    \cline{2-26}&Settings & \multicolumn{3}{c|}{Targeted} & \multicolumn{3}{c|}{Untargeted} & \multicolumn{3}{c|}{Targeted} & \multicolumn{3}{c|}{Untargeted} & \multicolumn{3}{c|}{Targeted} & \multicolumn{3}{c|}{Untargeted} & \multicolumn{3}{c|}{Targeted} & \multicolumn{3}{c}{Untargeted}  \\
    
    \cline{2-26}& Metrics & ASR   & Avg.Q&Med.Q & ASR   & Avg.Q &Med.Q& ASR   & Avg.Q &Med.Q& ASR   & Avg.Q &Med.Q& ASR   & Avg.Q &Med.Q& ASR   & Avg.Q &Med.Q& ASR   & Avg.Q&Med.Q & ASR   & Avg.Q&Med.Q  \\
    \hline

    \multirow{7}{1.8cm}{\centering Query  w/o \\surrogate} &     SignH& 28.1 &        9,021  &10,000& \underline{95.5} &           1,504 & 452& 60.8 &7,606  &8,401 & \textbf{100.0} &            702  &194 &49.8 &        7,935  & 9,950& \underline{99.5} &              931 & 212 &  45.7 &         8,315 & 10,000 & \textbf{100.0} &              363 &11  \\

   &  NES   & 37.7 &        8,748 & 10,000 & \textbf{100.0} &           1,234  & 682&47.7 &                     8,328 & 10,000 & 98.0 &         1,068 & 605 &54.3 &        7,988  &9,242 & 97.5 &           1,264  & 682&  45.7 &         8,235 & 10,000 & 91.0 &           1,711  &589 \\
  
    & $\mathcal{N}$Attack & 76.9 &        6,883  &7,323 & \textbf{100.0} &              878 & 606 & 93.5 & 5,924  & 5,858& \textbf{100.0} &            752  &505 & 94.5 &        5,324  & 5,101& \textbf{100.0} &              780 &606  & 90.0 &         5,644  &5,202 & \textbf{100.0} &              563 &404   \\

   &  SimBA &  77.5 &        5,874 & 5,489 & \textbf{100.0} &              865  & 419&\underline{97.5} &                     3,505 & 3,016 & 99.0 &            533  & 269& \underline{97.0} &        3,873  &3,445 & \underline{99.5} &              614 & 361 & 94.0 &         4,230  &3,620 & \underline{99.5} &              428 &  252\\

   &CGA   &90.9 &        4,645 & - & \textbf{100.0} &              \underline{139}  &- & 91.1 &                     4,801 &  -& \underline{99.4} &            137  & -& 91.6 &        4,222 &  -& 97.3 &              475  & -& 93.2 &         3,972  &- & 99.3 &              202 & -    \\
   
   & DifAttack & \underline{97.0} &       \underline{3,173}  & \underline{2,625}& \textbf{100.0} &  {165} & \underline{40} & \textbf{100.0} &      \underline{2,246}  &\underline{2,018} & \textbf{100.0} &  \underline{123}  & \underline{54}&\textbf{100.0} &       \underline{1,901} & \underline{1,793} & \textbf{100.0} & \underline{156}  & \underline{64}& \underline{98.0} &  \underline{2,793} & \underline{2,775} & \textbf{100.0} &  \underline{91}  & \underline{8} \\
 & Dif++$_{\text{open}}$ &\textbf{98.5} &\textbf{2,678}&\textbf{2,241}&\textbf{100.0}&\textbf{111}&\textbf{21}&\textbf{100.0}&\textbf{1,348}&\textbf{1,133}&\textbf{100.0}&\textbf{44}&\textbf{1}&\textbf{100.0}&\textbf{1,351}&\textbf{1,189}&\textbf{100.0}&\textbf{81}&\textbf{6}&\textbf{99.0}&\textbf{2,085}&\textbf{1,796}&\textbf{100.0}&\textbf{40}&\textbf{1} \\
 \hline
   \multirow{8}{1.8cm}{\centering Query  w/ \\surrogate}  &  BASES & 12.5 &        8,901  &10,000 & 81.7 &           2,080 &\textbf{1}  &  26.7 & 8,229  &10,000 & 95.5 &            564 & 69 & 15.4 &         9,228 & 10,000 & 85.1 &           1,717  & \textbf{1} &35.3 &        7,179  &10,000 & 86.4 &           1,549  & \textbf{1}   \\

  &   P-RGF  &60.2 &        5,222  & 4,588& 93.0 &           1,162 & 157 & 58.3 &                     6,093 & 7,043 & \underline{98.0} &            529  &85 &51.3 &        6,393 & 9,601 & \underline{96.0} &              711  &70 &  65.3 &         5,433 &4,995  & 95.0 &              796  & 119 \\

   & Subspace & 50.3 &        6,642 & 9,874 & 95.5 &              994  &40 & 68.3 &                     4,964  &3,984 & 95.5 &            867  & \underline{31}&66.3 &        4,750 & 3,034 & 94.5 &              917  &\underline{28} &  33.2 &         8,063   &10,000& 97.0 &              566   & \underline{25}\\
    
   &  GFCS  &  60.0 &        6,359 &7,101  & \underline{96.5} &           1,348  &463 &\underline{93.0} &                     3,358 & 2,494 & \textbf{100.0} &            515 & 107 &95.0 &        3,274 & 2,744 & \textbf{100.0} &              739  &229 & 72.5 &         6,127  &3,161 & \underline{98.5} &              658 & 128 \\

   &  MCGSquare   &\underline{96.0} & \underline{2,937} &\underline{2,132} &\textbf{100.0}  & \underline{174}     & \underline{14}      &\textbf{100.0} &   \underline{1,816}       &    \underline{1,568}         & \textbf{100.0} & \underline{49}     &\textbf{1}    & \underline{99.5}&\underline{1,346} &\underline{1,073}   & \textbf{100.0}& \underline{61}  &  \textbf{1}    &\underline{99.0} & \underline{1,633}  &\underline{1,323} & \textbf{100.0}&\underline{37}  & \textbf{1}                 \\

          &Dif++$_{\text{close}}$ &\textbf{99.5} &\textbf{1,020}&\textbf{541}&\textbf{100.0}&\textbf{19}&\textbf{1}&\textbf{100.0}&\textbf{432}&\textbf{229}&\textbf{100.0}&\textbf{11}&\textbf{1}&\textbf{100.0}&\textbf{573}&\textbf{369}&\textbf{100.0}&\textbf{20}&\textbf{1}&\textbf{99.5}&\textbf{859}&\textbf{506}&\textbf{100.0}&\textbf{27}&\textbf{1}  \\
     
    \hline\hline
    \end{tabular}%
    }
  \label{tab:simpleArch}%
\end{table*}%

\begin{table*}[th]
\small\tabcolsep=0.02cm
  \centering
  \caption{Results of score-based black-box attacks on ImageNet for the ``\textit{complex group}''.}
  \scalebox{0.75}{
    \begin{tabular}{c|c|ccc|ccc|ccc|ccc|ccc|ccc|ccc|ccc}
    \hline\hline
   \multirow{3}[2]{*}{Scenarios} & Victims & \multicolumn{6}{c|}{Swin-V2-T}  & \multicolumn{6}{c|}{EfficientNet-B3} & \multicolumn{6}{c|}{ResNet101} & \multicolumn{6}{c}{ConvNeXtBase} \\
 \cline{2-26}  & Settings & \multicolumn{3}{c|}{Targeted} & \multicolumn{3}{c|}{Untargeted} & \multicolumn{3}{c|}{Targeted} & \multicolumn{3}{c|}{Untargeted} & \multicolumn{3}{c|}{Targeted} & \multicolumn{3}{c|}{Untargeted} & \multicolumn{3}{c|}{Targeted} & \multicolumn{3}{c}{Untargeted}  \\
\cline{2-26}   & Metrics & ASR   & Avg.Q &Med.Q& ASR   & Avg.Q &Med.Q& ASR   & Avg.Q & Med.Q&ASR   & Avg.Q & Med.Q&ASR   & Avg.Q &Med.Q& ASR   & Avg.Q &Med.Q& ASR   & Avg.Q & Med.Q& ASR   & Avg.Q  &Med.Q\\
    \hline
   \multirow{7}{1.8cm}{\centering Query  w/o \\surrogate} &   SignH & 44.2 &      8,212 &10,000 & 82.9 &      3,049 & 968& 31.7 &        8,847 &10,000 & 87.4 &      2,611 &677 & 6.5 &        9,744 &10,000 & 66.3 &        4,670 &3,544 & 20.6 &        9,216& 10,000 & 81.9 &      3,979 &2,950 \\
  &  NES   & 30.7 &      8,722 & 10,000& 95.0 &      2,093 & 1,122& 15.6 &        9,382 & 10,000& 95.5 &      2,185 & 1,199& 10.6 &        9,606 & 10,000& 90.0 &        3,013 &1,913 & 11.1 &        9,471 & 10,000& 86.4 &      3,561& 2,499  \\
  &  $\mathcal{N}$Attack & 66.8 &      7,473&7,979  & \underline{99.0} &      1,298 &808 & 32.2 &        8,825 &10,000 & \underline{99.0} &      1,390  &808& 20.6 &        9,128& 10,000 & \underline{96.5} &        1,888  &1,263& 40.7 &        8,609 &10,000 & \textbf{100.0} &      2,012  &1,465  \\
   & SimBA & 84.0 &      5,822 &4,064 & 98.5 &      2,185 &990 & 65.0 &        7,585 &6,242 & 98.0 &      1,419  &584 &30.5 &        8,618 &10,000 & 73.0 &        3,581  &739& 90.0 &        4,809&4,249  & \underline{98.5} &      1,065& 406 \\
 &DifAttack & \underline{92.0} &      \underline{4,291}& \underline{3,729} & \textbf{100.0} &         \underline{409}& \underline{171} & \underline{69.0} &        \underline{6,080} &\underline{5,709} & \textbf{100.0} &         \underline{400} &\underline{151} & \underline{62.5} &       \underline{6,680}  &\underline{6,954}&  \textbf{99.0} &          \underline{523} &\textbf{149} & \underline{97.0} &        \underline{3,850}  &\underline{3,461}& \textbf{100.0} &         \underline{435}&  \underline{231} \\
           &  Dif++$_{\text{open}}$ & \textbf{{99.0}} & \textbf{{2,598} } &\textbf{{2,176}}& \textbf{100.0} & \textbf{170         } &\textbf{57}& \textbf{{89.0}} & \textbf{{4,818}}&\textbf{{4,224}} & \textbf{100.0} & \textbf{ 302        } &\textbf{129}& \textbf{{75.5}} & \textbf{ {5,946}      }& \textbf{{5,641}}& \textbf{{99.5}} & \textbf{  {482}        }&\underline{{161}} & \textbf{100.0} & \textbf{ {   1,706 }   } &\textbf{{1,456}}& \textbf{100.0} & \textbf{     149   }  &\textbf{1}\\
           
    \hline
  \multirow{6}{1.8cm}{
  \centering Query w/ \\ surrogate}   & P-RGF & 62.3 &      4,610 & 2,264& 96.5 &         755 & 66& 49.3 &        6,244& 10,000 & 90.0 &      1,345& 66 & 34.7 &        7,256& 10,000 & 89.5 &        1,301& 82 & 57.3 &        5,720 & 7,090& 88.9 &      1,381 &66 \\
   & Subspace & 61.3 &      6,597 & 7,252& \underline{99.0} &         882  &138& 55.3 &        6,834 &8,757 & \underline{95.5} &         993  &121 &35.7 &        7,978 & 10,000& 95.5 &        1,198& 136 & \underline{76.9} &        5,206 &4,297 & \underline{99.0} &         569 &135 \\
   &  GFCS  & 61.5 &      5,683 & 5,479& 86.5 &      1,794 & \underline{12} & 57.0 &        5,897 & 5,233& 89.0 &      1,692 &345 & 36.0 &        7,577& 10,000 & 87.0 &        1,521 &  \underline{14}& 71.0 &        4,270 & 2,690& 91.5 &      1,621  &  \underline{44}\\
     &  MCGSquare   & \underline{99.0}&        \underline{2,352}     &  \underline{1,934}        & \textbf{100.0} & \underline{251}  &   {72}   & \underline{95.0} & \underline{3,738} &\underline{3,255} & \underline{99.5} &  \underline{463} &  \underline{64}           &\underline{75.5} & \underline{6,079} &\underline{6,104}&\textbf{99.5}  & \underline{400}  & 15      &\textbf{100.0} & \underline{2,762}& \underline{2,429} &\textbf{100.0}& \underline{226} &53    \\
    
    &  Dif++$_{\text{close}}$ & \textbf{100.0} & \textbf{  294  } &\textbf{91}& \textbf{100.0} & \textbf{        60}& \textbf{1}& \textbf{98.5} & \textbf{   1,287     }&\textbf{766} & \textbf{100.0} & \textbf{   138      }& \textbf{1}& \textbf{94.0} & \textbf{   1,460     }&\textbf{226} & \textbf{99.5} & \textbf{   172       }& \textbf{1}& \textbf{100.0} & \textbf{   710   } &\textbf{301}& \textbf{100.0} & \textbf{   83     } & \textbf{1} \\
    \hline\hline
    \end{tabular}%
    }
  \label{tab:compliArch}%
\end{table*}%

\begin{table*}[t!]
  \centering
  \small
  \setlength{\tabcolsep}{1.6pt}
  \renewcommand{\arraystretch}{1.08}
  \caption{{Open-set attacks using different surrogate-model strengths. ``Strong'' denotes the surrogate set \{ResNet101, ConvNeXtBase, Swin-V2-T\}, while ``Weak'' denotes the surrogate set \{GoogLeNet, VGG16, SqueezeNet\}.}}
  \resizebox{\textwidth}{!}{
  \begin{tabular}{c|ccc|ccc|ccc|ccc|ccc|ccc}
    \hline\hline
    \multirow{3}{*}{Methods}
    & \multicolumn{6}{c|}{Food101}
    & \multicolumn{6}{c|}{ObjectNet}
    & \multicolumn{6}{c}{VGGFace2} \\
    \cline{2-19}
    & \multicolumn{3}{c|}{Targeted}
    & \multicolumn{3}{c|}{Untargeted}
    & \multicolumn{3}{c|}{Targeted}
    & \multicolumn{3}{c|}{Untargeted}
    & \multicolumn{3}{c|}{Targeted}
    & \multicolumn{3}{c}{Untargeted} \\
    \cline{2-19}
    & ASR & Avg.Q & Med.Q
    & ASR & Avg.Q & Med.Q
    & ASR & Avg.Q & Med.Q
    & ASR & Avg.Q & Med.Q
    & ASR & Avg.Q & Med.Q
    & ASR & Avg.Q & Med.Q \\
    \hline

    SimBA
    & 69.7 & 5,929 & 5,416 & 97.5 & 1,392 & 816
    & 18.0 & 9,210 & 10,000 & 60.0 & 5,474 & 5,809
    & 58 & 7,817 & 9,091 & \underline{99.5} & 1,417 & 1,075 \\

    SignH
    & 84.0 & 4,545 & 3,551 & \underline{99.0} & 745 & 261
    & 49.5 & 7,166 & 10,000 & 90.6 & 2,012 & 423
    & 23 & 9,109 & 10,000 & 97.5 & 1,755 & 1,159 \\

    $\mathcal{N}$Attack
    & 51.0 & 6,990 & 9,588 & \textbf{100.0} & 312 & 255
    & 36.0 & 7,585 & 10,000 & \textbf{100.0} & 556 & 217
    & 82 & 6,233 & 6,111 & \textbf{100} & 749 & 606 \\

    NES
    & 29.5 & 8,278 & 10,000 & \textbf{100.0} & 609 & 496
    & 22.0 & 8,597 & 10,000 & \underline{99.5} & 1,550 & 1,550
    & 22.5 & 9,112 & 10,000 & 99 & 1,704 & 1,326 \\

    DifAttack
    & 84.0 & 3,717 & 2,371 & \textbf{100.0} & 100 & \underline{41}
    & 46.5 & 6,693 & 10,000 & \textbf{100.0} & 270 & \textbf{11}
    & \textbf{100} & 2,520 & 2,166 & \textbf{100} & 191 & 156 \\
\hline
    Dif++$_{\text{open}}$ (Strong)
    & \textbf{90.0} & \textbf{2,942} & \textbf{1,466}
    & \textbf{100.0} & \textbf{59} & \textbf{31}
    & \textbf{57.5} & \textbf{5,825} & \textbf{5,551}
    & \textbf{100.0} & \textbf{204} & \underline{21}
    & \textbf{100} & \textbf{904} & \textbf{716} & \textbf{100} & \underline{66} & \textbf{51} \\

    {Dif++$_{\text{open}}$ (Weak)}
    & \underline{89.0} & \underline{2,966} & \underline{1,566}
    & \textbf{100.0} & \underline{68} & \textbf{31}
    & \underline{55.0} & \underline{5,854} & \underline{6,466}
    & \textbf{100.0} & \underline{216} & \textbf{11}
    & \textbf{100} & \underline{1,052} & \underline{861}
    & \textbf{100} & \textbf{65} & \textbf{51} \\

    \hline\hline
  \end{tabular}
  }
  \label{tab:attackclip}
\end{table*}

\section{Experiments}\label{sec:exp}
In this section, we conduct a comprehensive evaluation of {DifAttack++} against SOTA score-based black-box attack methods. The experiments encompass both \textit{closed-set} and \textit{open-set} scenarios, and further extend to real-world commercial APIs and defensive mechanisms. In addition, we analyze the adversarial sensitivity of the disentangled features and visualize their impact on image reconstruction. Finally, ablation studies are conducted to validate the contributions of the cross-domain architecture, the HDF module, and the initialization strategies.
\subsection{Experiment Setup}

\subsubsection{Datasets}
For closed-set scenarios, we utilize the large-scale \textit{ImageNet-1k}~\cite{deng2009imagenet}, hereafter referred to as ImageNet, as both the training and test dataset. Images are cropped and resized to $224\times224$ to match standard classifier inputs. For open-set scenarios, the training dataset remains ImageNet, whereas the test datasets differ and include \textit{Food101}~\cite{bossard2014food}, which contains 101 fine-grained food categories, and \textit{ObjectNet}~\cite{barbu2019objectnet}, consisting of natural images with complex rotations, backgrounds, and viewpoints. {We further include \textit{VGGFace2}~\cite{8373813}, a large-scale face recognition dataset, to evaluate the robustness of DifAttack++ under a larger semantic and domain gap from ImageNet.} Images are resized to $224\times224$ for Food101, $160\times160$ for VGGFace2, and $336\times336$ for ObjectNet to match the corresponding classifier inputs and maintain high clean accuracy. Notably, when attacking ObjectNet, we exclude 131 classes that overlap with ImageNet to ensure a strictly open-set evaluation, following the protocol in~\cite{barbu2019objectnet}.

\subsubsection{Victim and Surrogate Models}
We select a diverse suite of models to serve as victims and surrogates, categorized according to their architectures and training paradigms. 

\noindent{\textbf{Standard ImageNet Classifiers.} }
We collect eight pre-trained models with diverse architectures from Torchvision library \cite{Torchvision}. Since the official source code for ImageNet in the baseline method CGA~\cite{feng2022boosting} is unavailable, we evaluate our method on the same models used in CGA (ResNet18, VGG16, GoogleNet, and SqueezeNet) and report their official results for fair comparison. In addition, we conduct further experiments on higher-performance models, including ConvNeXtBase, EfficientNet-B3, Swin-V2-T, and ResNet101. Based on their clean classification accuracy on ImageNet, we partition these models into two groups: a ``simple'' group (clean accuracy $<80\%$) and a ``complex'' group (clean accuracy $\geq80\%$).
Following the protocol in CGA, we adopt a ``leave-one-out'' strategy within each group, where three models are used as surrogates to guide the attack against the remaining fourth victim model. 

\noindent{\textbf{Foundation Models.}} 
To further verify the efficacy against large-scale pre-training, and considering the pivotal role of CLIP in large vision-language models~\cite{zhang2024visually,zhang2024mm,tong2024cambrian,liu2024improved}, we target OpenAI’s CLIP models~\cite{ilharco_gabriel_2021_5143773}, including ViT-B/16 ( trained on
Web Image-Text pairs) and ViT-H-14-CLIPA-336 (trained on
the DataComp-1B~\cite{gadre2024datacomp} dataset). These models serve as victim classifiers in open-set evaluations on the Food101 and ObjectNet datasets. 

\noindent{\textbf{{Face Recognition Model.}}}
{For the open-set evaluation on VGGFace2, we use the InceptionResnetV1 face-recognition model pretrained on VGGFace2 as the victim classifier~\cite{facenet}}.

\noindent{\textbf{Real-world API \& Defenses.} }
We also extend our evaluation to the commercial Imagga API~\cite{Imagga}, as well as to a range of defensive mechanisms. The specific configurations and settings for these targets are detailed in the corresponding subsections.

\subsubsection{Comparative Methods}
For a fair comparison, based on the reliance on surrogate models during the attack stage, we divide the comparative methods into two groups: 
1) \textit{surrogate-free} methods, including SignH \cite{al2020sign}, NES \cite{ilyas2018black}, $\mathcal{N}$Attack \cite{li2019nattack}, SimBA \cite{guo2019simple}, CGA \cite{feng2022boosting}, DifAttack \cite{JunDifAttack2024}, and our proposed \textbf{$\text{Dif++}_{\text{open}}$} (which uses Gaussian initialization without TI); 
and 2) \textit{surrogate-assisted} methods, including BASES \cite{cai2022blackbox}, P-RGF \cite{cheng2019improving}, Subspace \cite{guo2019subspace}, GFCS \cite{lord2022attacking}, MCGSquare (a combination of MCG \cite{yin2023generalizable} and Square \cite{andriushchenko2020square}, representing the SOTA in ASR and query efficiency), as well as our \textbf{$\text{Dif++}_{\text{close}}$} (which initializes the AF from the image augmented with transferable perturbations).

\subsubsection{Implementation Details}
Consistent with the baselines, the maximum perturbation budget is set to $\epsilon=12/255$ under the $l_\infty$ norm, while the maximum query budget is set to $10{,}000$ for offline models. 
For the real-world Imagga API, the query budget is further restricted to $500$ in order to reflect practical cost constraints. 
Regarding the hyperparameters of DifAttack++, we set the sampling variance $\sigma=0.1$, the downscaling factor $s=4$, the learning rate $\eta=0.01$, and the margin constant $\kappa=5$. All experiments are conducted on NVIDIA GeForce RTX 3090 GPUs. {The autoencoders are trained for one epoch on ImageNet with batch size $16$, taking approximately $29$ hours on a single GPU. This offline training is typically performed once for a given surrogate set before attacking. The peak training memory depends on the surrogate models: it is about $7.4$--$13.2$ GB when using three simple surrogate models and $19.4$--$23.0$ GB when using three complex surrogate models. During the attack stage, the memory usage is much lower, typically around $1$--$2$ GB.}

\begin{table}[t]
  \centering
  \caption{Results of surrogate-free score-based black-box attacks against real-world Imagga API in open-set scenarios.}
  \scalebox{1}{
    \begin{tabular}{c|ccc|ccc}
    \hline\hline
    Settings$\rightarrow$ & \multicolumn{3}{c|}{Untargeted} & \multicolumn{3}{c}{Targeted} \\
    \hline
    \multicolumn{1}{c|}{Methods$\downarrow$} & {ASR} & {Avg.Q}& {Med.Q} & {ASR} & \multicolumn{1}{c}{Avg.Q}& \multicolumn{1}{c}{Med.Q} \\
    \hline
    SignH & {70.0} & {186} &\underline{14}& {63.6} & \multicolumn{1}{c}{212} &\textbf{10}\\
    $\mathcal{N}$attack & {76.7} & {192} &31& {38.2} & \multicolumn{1}{c}{406} &61\\
    SimBA & {60.0} & {292} & 364&{54.5} & \multicolumn{1}{c}{244}& 261\\
   
    DifAttack & \underline{80.0} & \underline{132} &25& \underline{68.2} & \underline{197}&30 \\
    Dif++$_{\text{open}}$ &   \textbf{86.7}    &  \textbf{92}&     \textbf{6} & \textbf{72.7}      & \textbf{163}&\underline{21}\\
    \hline\hline
    \end{tabular}%
     }
  \label{tab:imagga}%
\end{table}%

\begin{table}[t]
\small\tabcolsep=0.06cm
  \centering
  \caption{Results of comparative methods against various SOTA defensive techniques on ImageNet in untargeted settings.}
  \scalebox{0.95}{
    \begin{tabular}{c|c|c|cccc}
    \hline\hline
    Scenarios&Methods & AT    & BitRed & NRP   & DiffPure & Avg.ASR \\
    \hline
  \multirow{6}{1.8cm}{\centering Query  w/o \\surrogate}&  SignH & 36.7  & 38.0    & 28.9  & 12.6  & 29.1 \\
    &NES   & 18.6  & 21.0    & 31.3  & 20.0    & 22.7 \\
    & $\mathcal{N}$Attack & 39.2  & 45.0    & 34.9  & 24.0    & 35.8 \\
    & SimBA & 39.5  & 14.5  & 37.2  & 13.1  & 26.0 \\
      &  DifAttack & \underline{52.5}  & \underline{46.5}  & \underline{41.5}  & \underline{28.5}  & \underline{42.3} \\
  &   Dif++$_{\text{open}}$ & \textbf{64.5} & \textbf{88.8} & \textbf{45.4} & \textbf{40.4} & \textbf{59.8} \\
      \hline
  
  \multirow{6}{1.8cm}{\centering Query  w/ \\surrogate}   & BASES & 0.0    & 58.5  & 32.6  & 8.6   & 24.9 \\
   & P-RGF & 43.2  & 38.0   & \underline{38.5}  & 30.4  & 37.5 \\
    &Subspace & 54.8  & 54.0    & 31.2  & 14.8  & 38.7 \\
   & GFCS  & 6.8   & 21.5  & 32.9  & 22.5  & 20.9 \\
   & MCGSquare  & \underline{70.5} &\underline{72.0} &34.5 &\underline{65.5}& \underline{60.6} \\

  &  Dif++$_{\text{close}}$ & \textbf{71.0} & \textbf{92.0} & \textbf{43.5} & \textbf{69.5} & \textbf{69.0} \\
    \hline\hline
    \end{tabular}%
    }
  \label{tab:defense}%
\end{table}%

\begin{table}[t!]
  \centering
  \small\tabcolsep=0.06cm
  \caption{{Average attack performance and perceptual quality of the generated AEs over eight ``Complex" and ``Simple" victim models and both targeted/untargeted settings on the ImageNet test dataset.}}
  \scalebox{0.8}{
    \begin{tabular}{c|c|cccccc}
    \hline\hline
    Scenarios & \multicolumn{1}{l|}{Method} & \multicolumn{1}{l}{ASR} & \multicolumn{1}{l}{Avg.Q} & \multicolumn{1}{l}{Med.Q} & \multicolumn{1}{l}{PSNR$\uparrow$} & \multicolumn{1}{l}{SSIM$\uparrow$} & LPIPS$\downarrow$ \\
    \hline
    \multirow{7}{1.8cm}{Query w/o \\surrogate} & SignH &                   62.6  &                5,419  &                5,460  &                26.27  &             0.6191  &             0.1906  \\
          & NES   &                   62.9  &                5,413  &                5,533  &                36.46  &             0.9043  &             0.0300  \\
          & $\mathcal{N}$Attack &                   81.9  &                4,211  &                4,246  &                35.83  &             0.9141  &             0.0342  \\
          & SimBA &                   87.6  &                3,438  &                2,759  &                37.56  &             0.9273  &             0.0443  \\
\cline{2-8}          & DifAttack &                   94.7  &                2,082  &                1,862  &                27.05  & \textbf{            0.8861 } & \textbf{            0.0691 } \\
          & Dif++$_\text{open}$ & \textbf{                  97.5 } & \textbf{               1,477 } & \textbf{               1,247 } & \textbf{               28.51 } &             0.7661  &             0.2092  \\
\hline \multirow{6}{1.8cm}{Query w/ \\ surrogate} & BASES &                   54.8  &                4,931  &                5,009  &                26.21  &             0.6075  &             0.3255  \\
          & P-RGF  &                   74.1  &                3,434  &                3,518  &                39.11  &             0.9337  &             0.0418  \\
          & Subspace &                   76.2  &                3,626  &                3,616  &                32.24  &             0.7931  &             0.1713  \\
          & GFCS  &                   80.9  &                3,277  &                2,515  &                43.60  &             0.9723  &             0.0280  \\
\cline{2-8}          & MCGSquare &                   97.7  &                1,520  &                1,252  &                26.23  &             0.8166  &             0.2780  \\
          & Dif++$_\text{close}$ & \textbf{                  99.3 } & \textbf{                    471 } & \textbf{                    199 } & \textbf{               29.11 } & \textbf{            0.8256 } & \textbf{            0.1560 } \\
    \hline\hline
    \end{tabular}%
  }
  \label{tab:visualquality}%
\end{table}%

\subsection{Performance in Closed-Set Scenarios}

In closed-set scenarios, the attacker has access to surrogate models trained on the same dataset as the victim model, making both surrogate-free and surrogate-assisted methods applicable. Therefore, we report the performance of both types of methods under this setting. Tables~\ref{tab:simpleArch} and \ref{tab:compliArch} summarize the attack performance in terms of Attack Success Rate (ASR), average queries (Avg.Q), and median queries (Med.Q) on victim models from the ``simple group'' and ``complex group'', respectively. Across extensive experimental settings, both variants of DifAttack++, namely Dif++$_{\text{close}}$ and Dif++$_{\text{open}}$, consistently achieve higher ASR and substantially lower query costs than their corresponding baselines.

\noindent\textbf{Effectiveness of Zero-Shot AE Check.}
In untargeted attacks, Dif++$_{\text{close}}$ achieves a Med.Q of $1$ across almost all victim models while maintaining equal or higher ASR. Med.Q$=1$ indicates that the candidate AE generated by the zero-shot AE check already succeeds in more than half of the cases before iterative query refinement. {This early success benefits from both the adversarial TI and its reconstruction through the learned AF/VF space. Therefore, the zero-shot AE check can be viewed as an effective early-verification mechanism that exploits the compatibility between TI and our disentangled reconstruction space. In contrast, the runner-up method MCGSquare attains $\text{Med.Q}=1$ only on victim models in the simple group; when attacking complex architectures, its Med.Q rises sharply, typically exceeding $50$. These results show that Dif++$_{\text{close}}$ can substantially reduce early query costs even against strong and modern architectures.} 

{For Dif++${_\text{open}}$, no transferable AE is used for initialization. Instead, the initial query is generated from the reconstructed clean image. Therefore, the low Med.Q of Dif++${_\text{open}}$ provides evidence that the learned AF space itself contains adversarially sensitive directions. For example, as shown in Tables~I and II, Dif++$_{\text{open}}$ achieves Med.Q = 1 on VGG16, SqueezeNet, and ConvNeXtBase. This suggests that, in these cases, a very local search around the clean AF feature is already sufficient to find successful AEs.} 

\noindent\textbf{Targeted Attacks.}
In the more challenging targeted setting, Dif++$_{\text{close}}$ maintains a clear performance lead in both effectiveness and efficiency. For example, when attacking ResNet101, our method improves ASR by $18.5\%$ over MCGSquare while simultaneously reducing the Med.Q by $96\%$. This substantial reduction in query cost highlights the practical efficiency of our proposed strategy under challenging targeted constraints.

\noindent\textbf{Generalization Across Architectures.}
DifAttack++ maintains a stable performance advantage even when the surrogate and victim models belong to substantially different architectural families—such as CNNs (e.g., ResNet, EfficientNet), hierarchical vision transformers (e.g., Swin-V2-T), and modern convolutional backbones inspired by transformers (e.g., ConvNeXt). This result confirms that our cross-domain disentanglement framework captures generalized adversarial features that are robust to architectural discrepancies, rather than merely overfitting to surrogate-specific gradients.

\noindent\textbf{Superiority of Dif++$_{\text{open}}$.} Furthermore, even in the absence of surrogate models, Dif++$_{\text{open}}$ outperforms most surrogate-assisted baselines in terms of both ASR and query efficiency. Notably, it achieves a query efficiency gain of up to 38.2\% compared to MCGSquare in targeted attacks against ConvNeXtBase. This superiority is consistently observed in targeted attack scenarios on GoogleNet, VGG16,
ResNet18, ConvNeXtBase, and ResNet101, as well as in untargeted attack
scenarios on GoogleNet, VGG16, Swin-V2-T, EfficientNet-B3,
and ConvNeXtBase. Compared with the vanilla DifAttack, Dif++$_{\text{open}}$ demonstrates substantial performance improvements, which further validates the effectiveness of our proposed decoupling mechanism and refined attack strategy. The individual contributions of these modules are elaborated in the subsequent ablation studies.

\subsection{Performance in Open-Set Scenarios}
\label{sec:exp:openset}

In this subsection, we evaluate the attack performance under the more challenging open-set scenarios. This setup tests the generalization capability of attack methods under domain shifts. For these experiments, we employ methods from the surrogate-free group.

\noindent\textbf{CLIP.} 
Table~\ref{tab:attackclip} presents the comparative results against CLIP models {under different open-set conditions. Dif++$_{\text{open}}$ consistently improves attack performance over baseline methods on Food101 and ObjectNet. In particular, when trained with the Strong surrogate set \{ResNet101, ConvNeXtBase, Swin-V2-T\}, Dif++$_{\text{open}}$ achieves the best targeted ASR and query efficiency on both Food101 and ObjectNet.} For example, on ObjectNet, it improves the targeted ASR from 46.5\% to 57.5\% over DifAttack and reduces Avg.Q from 6,693 to 5,825. In the untargeted setting, it also maintains 100.0\% ASR with substantially fewer queries. {We further evaluate the robustness to weaker surrogate models by training Dif++$_{\text{open}}$ with the Weak surrogate set \{GoogLeNet, VGG16, SqueezeNet\}. The weak-surrogate version is slightly worse than the strong-surrogate version in most targeted settings, e.g., targeted ASR decreases from 90.0\% to 89.0\% on Food101 and from 57.5\% to 55.0\% on ObjectNet. Nevertheless, it still clearly outperforms DifAttack in both ASR and query efficiency, showing that the learned AF/VF space does not heavily rely on very strong surrogate models.}

These results suggest that the proposed cross-domain disentanglement strategy learns general AFs that can transfer across different target domains. Even without prior knowledge of the target domain, optimizing in the learned adversarial latent space yields stronger open-set performance than optimizing directly in the pixel space, as in NES, or in the $\tanh$-transformed pixel space, as in $\mathcal{N}$Attack, although these methods all rely on natural evolution strategy for query-based optimization.

\noindent\textbf{Face Images.} 
{To evaluate a larger domain gap, we conduct an additional open-set experiment on VGGFace2, where the autoencoders are still trained with ImageNet-based surrogate models but the target domain consists of face images. Even under this substantial semantic shift, Dif++$_{\text{open}}$ remains effective. With Weak surrogate models, it maintains 100.0\% ASR in both targeted and untargeted attacks, while reducing targeted Avg.Q/Med.Q from 2,520/2,166 to 1,052/861 compared with DifAttack. In the untargeted setting, it further reduces Avg.Q/Med.Q from 191/156 to 65/51. The Strong surrogate version achieves even lower targeted query counts, reducing Avg.Q/Med.Q to 904/716. These results show that Dif++$_{\text{open}}$ can generalize to a substantially different target domain, although performance may still degrade when the domain gap becomes too large or when the available surrogate models are too weak.}

\noindent\textbf{Real-World API.} 
To further assess practical applicability, we conduct attacks against the commercial Imagga API \cite{Imagga}. This API employs a recognition model trained on a massive and undisclosed dataset covering over $3,000$ categories, thereby representing a true ``black-box'' setting with unknown training data. Due to API call limitations, we compare against several top-performing open-set methods, including DifAttack, SimBA, $\mathcal{N}$Attack, and SignH. For untargeted attacks, the objective is to remove the top-3 predicted classes, whereas for targeted attacks, we aim to promote the second-highest scoring class to the top rank. As shown in Table \ref{tab:imagga}, $\text{Dif++}_{\text{open}}$ consistently dominates in this real-world scenario.
Despite the complete lack of knowledge regarding the API’s training distribution, our method reliably achieves higher ASR with fewer queries.
In particular, for targeted attacks, our approach significantly reduces the query cost compared to query-intensive methods such as SimBA and $\mathcal{N}$Attack. These results indicate that the proposed disentangled feature space remains effective even in a large-scale and unknown label space. Overall, this experiment strongly underscores the practical utility and cost-effectiveness of DifAttack++ for evaluating the robustness of commercial AI services.

\subsection{Robustness against Defensive Mechanisms}
\label{sec:exp:defense}

To verify the resilience of the generated AEs, we evaluate the performance of \text{DifAttack++} against SOTA defensive strategies. These defenses can be broadly categorized into two types: model-level defenses, such as Adversarial Training (AT) \cite{salman2020adversarially}, and input-level purification methods, including Bit-depth Reduction (Bit-Red) \cite{featureSqueeze}, Neural Representation Purifier (NRP) \cite{naseer2020self}, and Diffusion Purifier (DiffPure) \cite{nie2022diffusion}. For the {AT} defense, following the protocol in CGA \cite{feng2022boosting}, we employ an adversarially trained ResNet-50 as the surrogate model. Subsequently, all methods are evaluated by generating AEs to attack a more robust, adversarially trained Wide-ResNet-50-2 victim model.
In contrast, for input purification defenses, we generate AEs targeting a standard ResNet-18 model on ImageNet. These AEs are then passed through the corresponding purification modules before being classified by the victim model.

The comparative results in Table \ref{tab:defense} clearly demonstrate the superior robustness of our method. DifAttack++ achieves the highest average ASR across all four defensive techniques and under both attack settings (query with or without surrogate models). Notably, Dif++$_\text{close}$ surpasses the runner-up method, MCGSquare, by a substantial margin of $8.4\%$ in terms of the averaged ASR across different defensive techniques. Specifically,
the resilience to input-level purification techniques suggests that our cross-domain fusion strategy effectively embeds adversarial features into the semantic structure of the image. In contrast to pixel-level noise, which can be easily smoothed out, the disentangled adversarial features fused by our HDF module are more integral to the image representation, and thus difficult to purify without compromising visual fidelity.
Furthermore, the success against the AT-protected victim model further confirms that our disentanglement mechanism is capable of extracting robust and generalized adversarial features even from robust surrogate models, thereby validating the overall efficacy of the proposed framework.

\begin{figure*}[t!]
\centering
\includegraphics[width=0.98\textwidth]{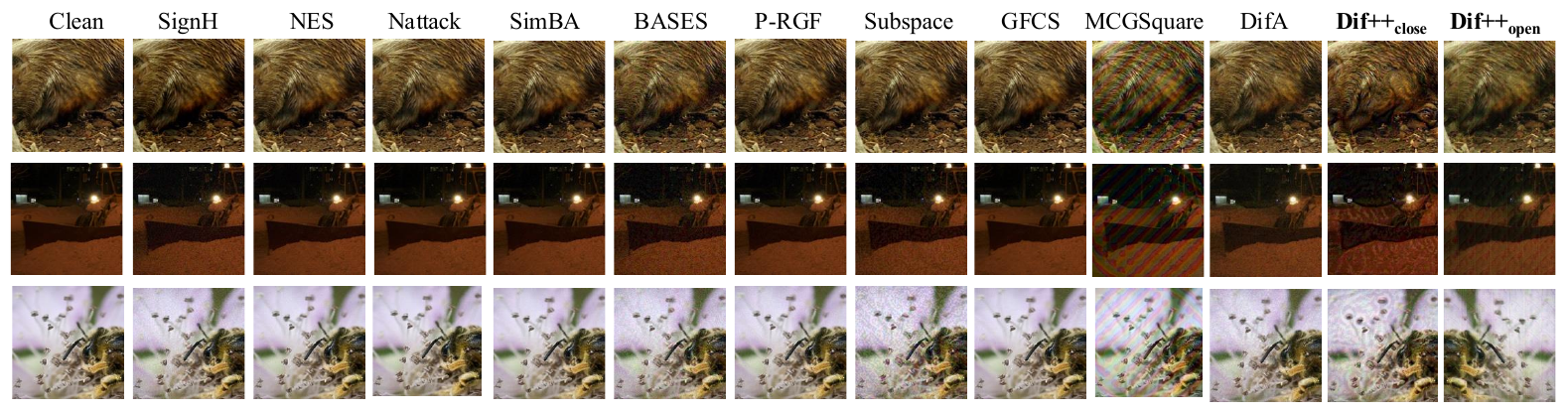} 
\caption{Visualization of AEs generated by SOTA score-based attack methods. In the surrogate-assisted setting, Dif++$_{\text{close}}$ produces visually subtler AEs with fewer structured artifacts than MCGSquare. In the surrogate-free setting, Dif++$_{\text{open}}$ and DifAttack show comparable visual appearance in the displayed examples.}
\label{fig:visae}
\end{figure*}

\begin{figure*}[t!]
\centering
\includegraphics[width=1\textwidth]{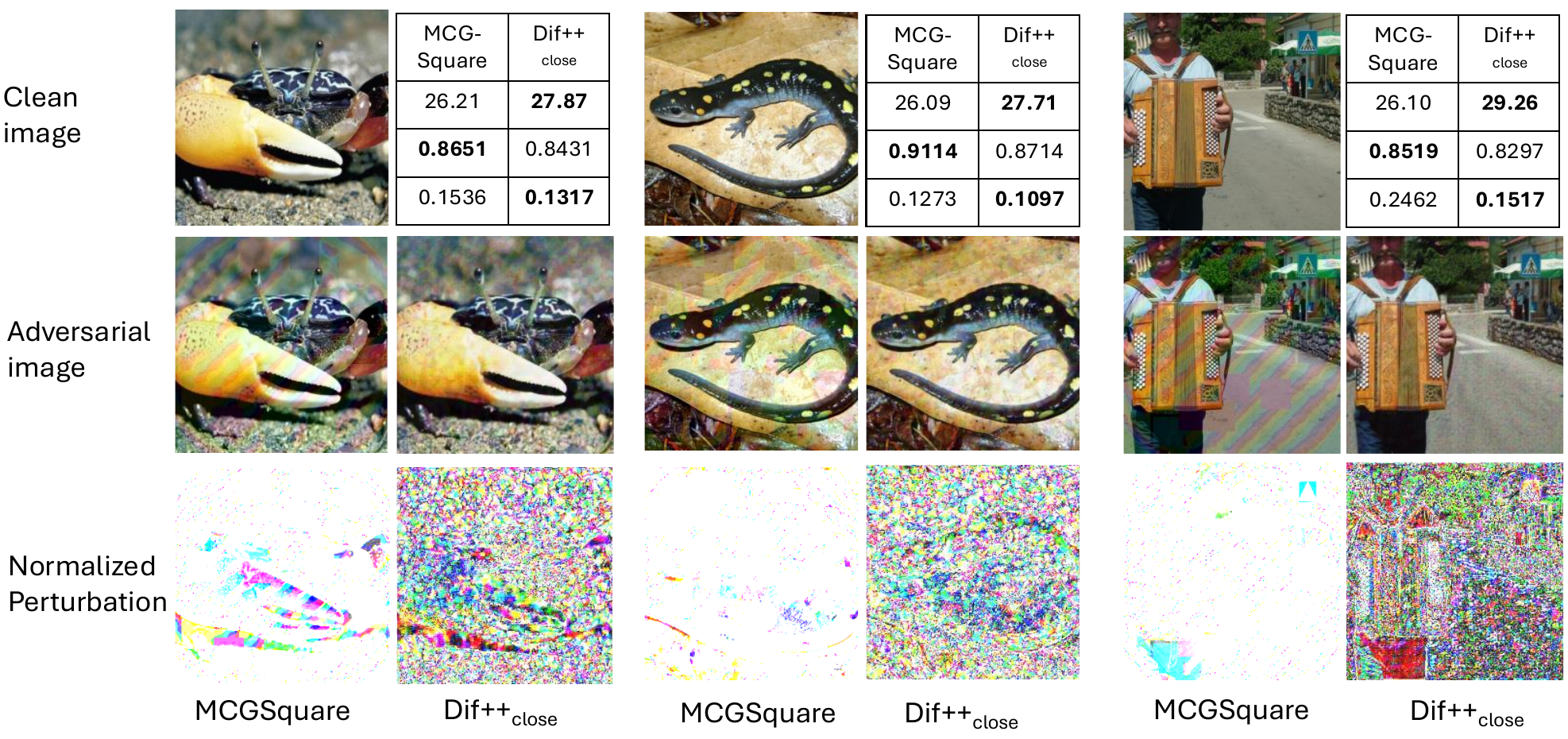}
\caption{
{Visualization of representative rare visual-quality trade-off cases in the closed-set scenario. Perturbation maps are independently normalized for visualization.
These AEs generated by Dif++$_{\text{close}}$ obtain higher PSNR and lower LPIPS, but lower SSIM, than those generated by the runner-up method, MCGSquare. These examples show that different perceptual metrics may not always improve consistently across individual samples.}
}
\label{fig:failClose}
\end{figure*}

\subsection{{Perceptual Quality Evaluation}}
\label{sec:exp:visualquality}

{We first provide a qualitative comparison of the generated AEs in Fig.~\ref{fig:visae}. In the surrogate-assisted setting, Dif++$_{\text{close}}$ produces visually subtler AEs with fewer structured artifacts than the corresponding runner-up, MCGSquare. In the surrogate-free setting, Dif++$_{\text{open}}$ and DifAttack show comparable visual appearance in the displayed examples, suggesting that visual differences are less obvious by inspection alone.}

{To further quantify perceptual distortion, we report PSNR, SSIM, and LPIPS in Table~\ref{tab:visualquality} under the same $\ell_\infty$ constraint. For clarity and space efficiency, Table~\ref{tab:visualquality} reports the average results over eight ``Complex'' and ``Simple'' victim models and both targeted and untargeted settings. The results show that DifAttack++ achieves a favorable trade-off between attack effectiveness, query efficiency, and perceptual distortion. In the surrogate-assisted setting, Dif++$_{\text{close}}$ not only improves ASR and query efficiency over MCGSquare, but also improves PSNR from 26.23 to 29.11, improves SSIM from 0.8166 to 0.8256, and reduces LPIPS from 0.2780 to 0.1560. In the surrogate-free setting, Dif++$_{\text{open}}$ improves ASR and query efficiency over DifAttack and also achieves higher PSNR. However, its SSIM and LPIPS are less favorable than those of DifAttack, indicating that different perceptual metrics may not always improve consistently.}

\begin{figure*}[t!]
\centering
\includegraphics[width=1\textwidth]{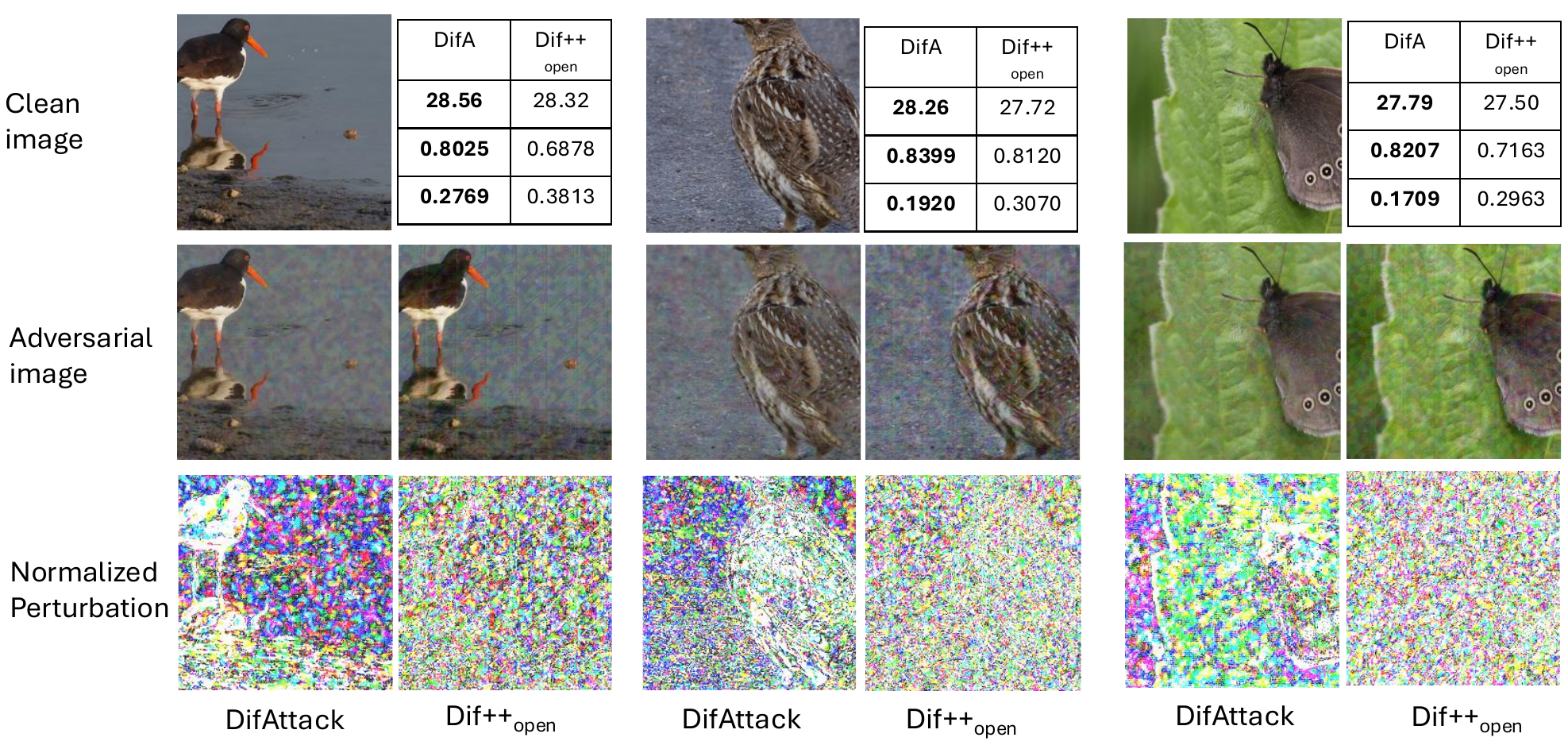}
\caption{
{Visualization of the only three visual-quality trade-off cases observed among the 200 open-set test images. Perturbation maps are independently normalized for visualization. These AEs generated by Dif++$_{\text{open}}$ obtain lower PSNR/SSIM and higher LPIPS than those generated by the runner-up method, DifAttack. The corresponding perturbations produced by Dif++$_{\text{open}}$ are more spatially distributed than those in the higher-PSNR cases shown in Fig.~\ref{fig:succ}, where Dif++$_{\text{open}}$ achieves higher PSNR than DifAttack. This suggests that more distributed perturbations may lead to less favorable perceptual metrics in some individual samples, although the attacks are still successful.
}
}
\label{fig:fail}
\end{figure*}

\begin{figure*}[t!]
\centering
\includegraphics[width=1\textwidth]{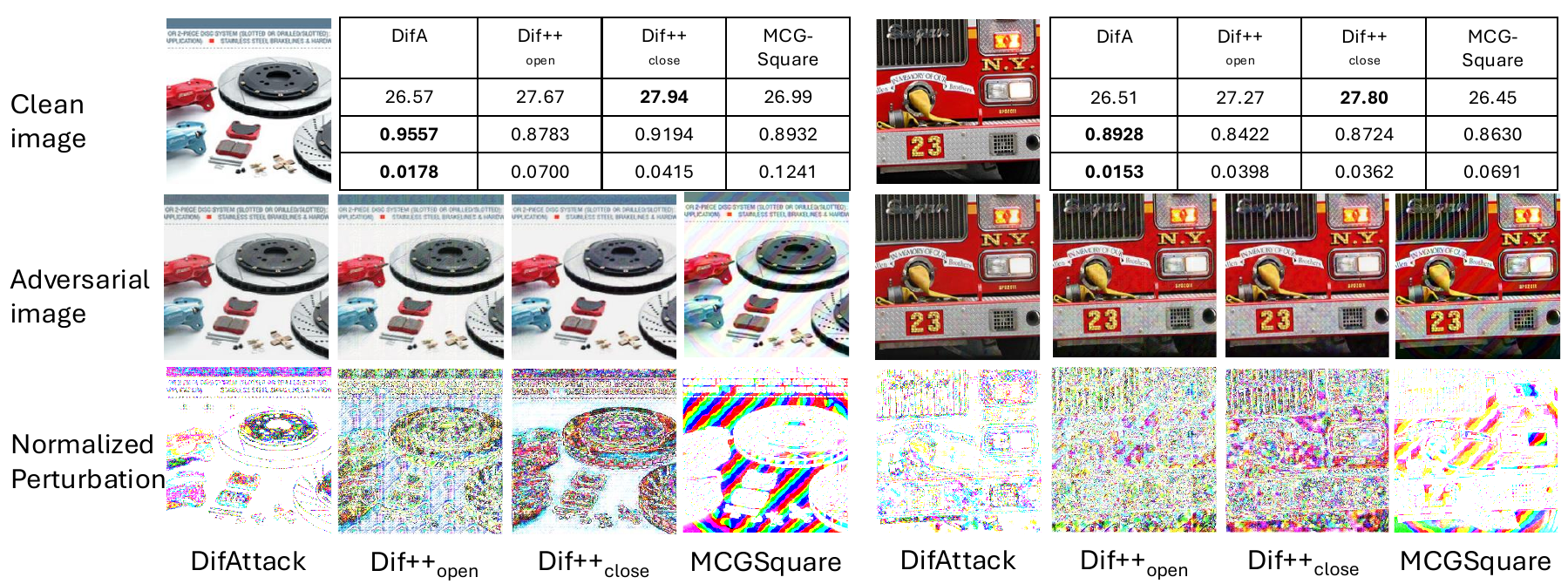} 
\caption{
{Visualization of representative typical visual-quality cases. Perturbation maps are independently normalized for visualization. In the closed-set scenario, the AEs generated by Dif++$_{\text{close}}$ obtain higher PSNR/SSIM and lower LPIPS than those generated by the runner-up method, MCGSquare. 
In the open-set scenario, the AEs generated by Dif++$_{\text{open}}$ obtain higher PSNR, but lower SSIM and higher LPIPS, than those generated by the runner-up method, DifAttack. 
These examples complement the rare trade-off cases in Fig.~\ref{fig:failClose} and Fig.~\ref{fig:fail}.}
}
\label{fig:succ}
\end{figure*}

{Since these averaged metrics may hide sample-level variations, we provide representative visual-quality examples in Figs.~\ref{fig:failClose}-\ref{fig:succ}, including both typical favorable cases and a small number of trade-off cases. These examples show that visual quality can vary across individual samples, and PSNR, SSIM, and LPIPS may not always change consistently. In the open-set scenario, Fig.~\ref{fig:fail} shows representative trade-off cases where Dif++$_{\text{open}}$ obtains lower PSNR/SSIM and higher LPIPS than DifAttack. The normalized perturbation maps indicate that the perturbations produced by Dif++$_{\text{open}}$ are more spatially distributed in these samples, which may explain the less favorable perceptual metrics. In contrast, Fig.~\ref{fig:succ} shows typical favorable cases where Dif++$_{\text{open}}$ achieves higher PSNR than DifAttack. In the closed-set scenario, Fig.~\ref{fig:failClose} shows representative trade-off cases against MCGSquare, where Dif++$_{\text{close}}$ may show slightly less favorable SSIM values for individual samples, despite its better average perceptual quality and much stronger attack efficiency in Table~\ref{tab:visualquality}.}

{Overall, Fig.~\ref{fig:visae}, Table~\ref{tab:visualquality}, and the sample-level examples in Figs.~\ref{fig:failClose}--\ref{fig:succ} show that DifAttack++ achieves a favorable balance among attack effectiveness, query efficiency, and perceptual quality. Although a small number of samples or some baseline methods may obtain better values on individual perceptual metrics, these cases usually come with lower ASR or higher query costs.}

 \begin{figure}[t!]
\centering
\includegraphics[width=0.48\textwidth]{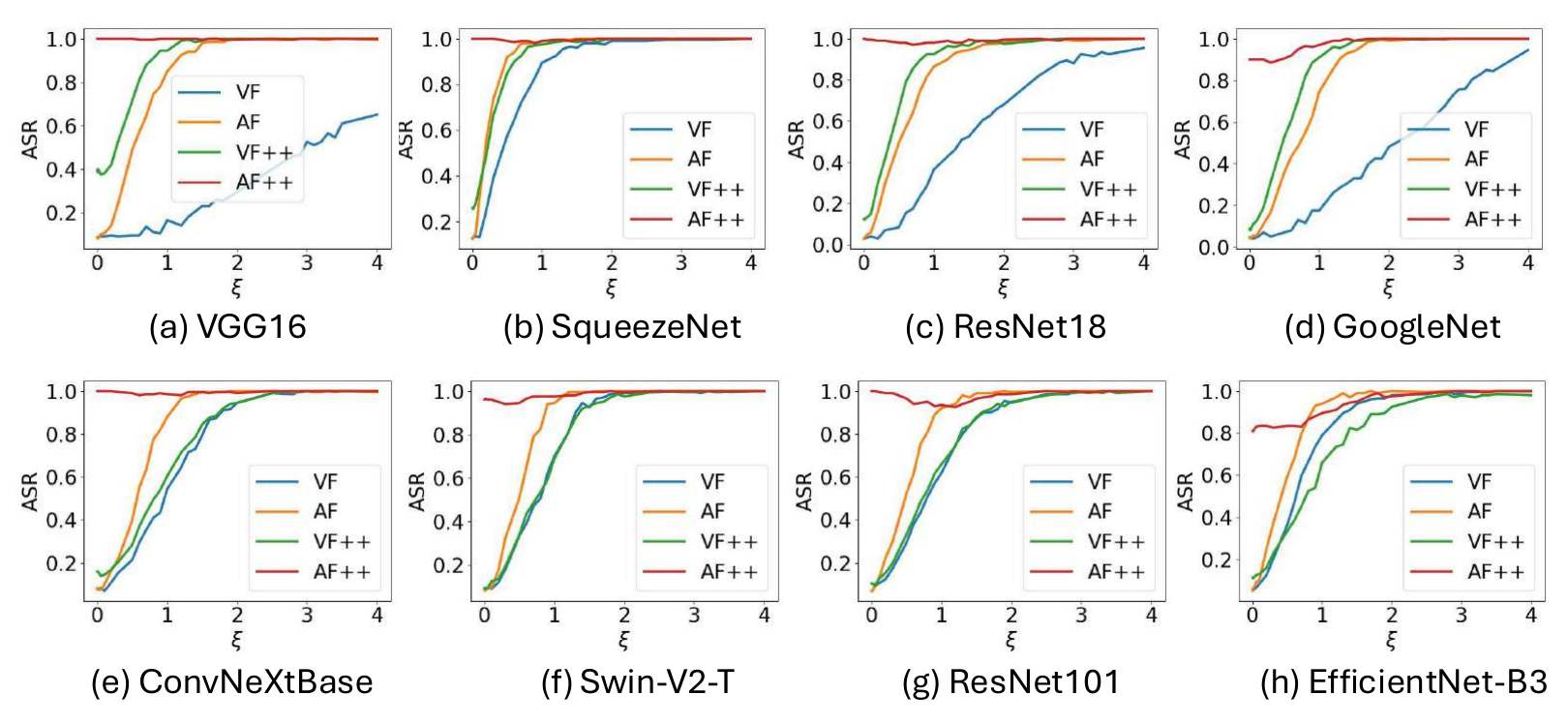} 
\caption{Adversarial sensitivity of clean images' VF and AF from DifAttack, or VF++ and AF++ from DifAttack++.
The x-axis starts at 0.001. Higher ASR values indicate greater adversarial sensitivity.}
\label{fig:featsens}
\end{figure}

\begin{figure}[t!]
	\centering
	\includegraphics[width=0.45\textwidth]{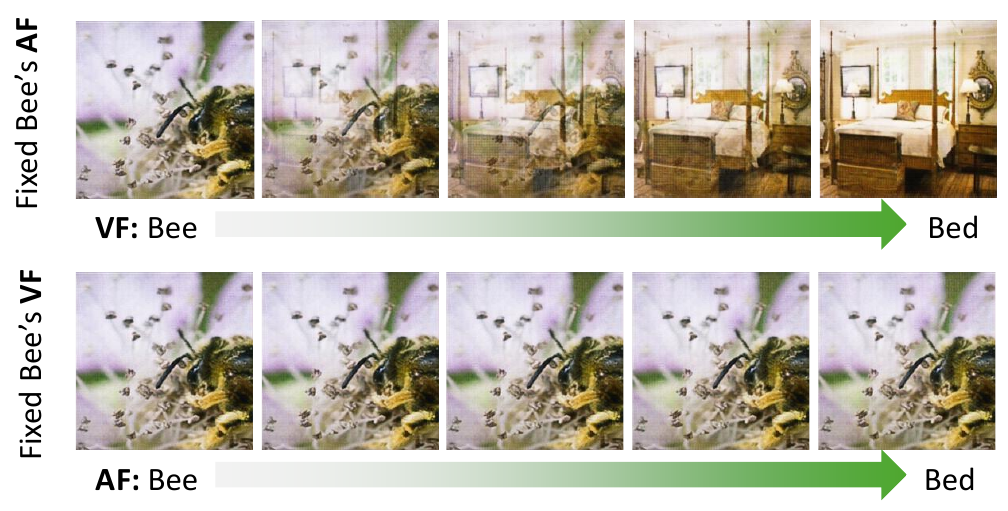}
	\caption{Progressive fusion of clean images' disentangled features extracted by the autoencoder $\mathcal{G}$ in DifAttack++.
Increasing VFs' contribution induces a gradual visual transition between images, whereas modifying AFs with fixed VFs produces negligible perceptual changes, highlighting the distinct roles of VFs and AFs.} 
	\label{fig:featurevis_detail} %
\end{figure}

\begin{figure*}[t!]
	\centering
	\includegraphics[width=0.95\textwidth]{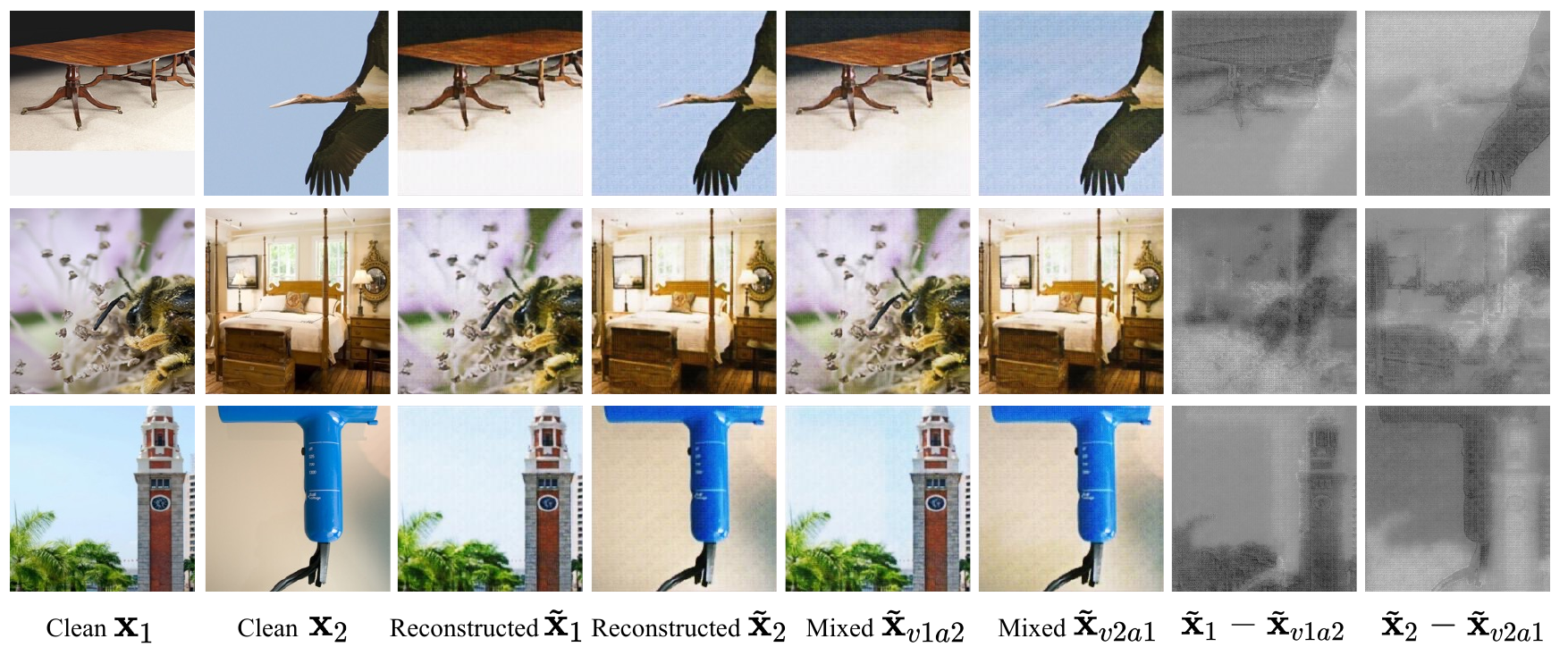}
\caption{Visualization of clean images' disentangled representations extracted by the autoencoder $\mathcal{G}$ in DifAttack++.
The reconstructed images show that the overall visual appearance is mainly preserved by the VF, whereas swapping the AF introduces localized semantic changes concentrated on foreground objects. This suggests that AF captures class-discriminative attention regions while {causing only subtle visual changes under the reconstruction process.}}
	\label{fig:featurevis} %
\end{figure*}

\subsection{Analysis on Disentangled Features}
\label{sec:exp:analysis}

To verify the disentanglement efficacy of {DifAttack++} and to elucidate the distinct roles of VF and AF, we conduct both quantitative sensitivity analyses and qualitative visualization experiments on ImageNet.

\noindent\textbf{Quantitative Analysis on Adversarial Sensitivity.}
We first examine the impact of perturbing specific features on attack performance. Specifically, we perform an ablation study in which Gaussian noise with varying standard deviation $\xi$ is injected exclusively into either VFs or AFs of clean images, extracted via the single-encoder in DifAttack (baseline) or the cross-domain autoencoders in DifAttack++, while keeping the counterpart feature fixed. The reconstructed images are then evaluated across eight diverse classifiers. As evidenced by the results illustrated in Fig.~\ref{fig:featsens}, the AF extracted by the autoencoder $\mathcal{G}^*$ in DifAttack++ (AF++) exhibits markedly superior adversarial sensitivity. In particular, the ASR curve of AF++ (red line) behaves like a step function, saturating to nearly $100\%$ almost immediately, even under negligible perturbation (e.g., $\xi=0.001$). In sharp contrast, the baseline AF (orange line) shows a much slower, sigmoid-like increase, requiring significantly larger perturbation magnitudes to achieve comparable success. 
Conversely, perturbing the VF (blue and green lines) leads to much smaller changes in ASR, which confirms that the VF is largely decoupled from adversarial effectiveness. In a few cases (e.g., Figs.~\ref{fig:featsens} (g) and (h)), the ASR of AF++ marginally lags behind that of AF at high perturbation levels ($\xi > 0.05$). However, this discrepancy is of limited practical relevance. Since pixel-domain distortion is not clipped in this sensitivity analysis, such relatively large $\xi$ values typically exceed the predefined perturbation constraint in practice. As a result, perturbations with $\xi < 0.05$ are more meaningful for guiding our attack method. Consequently, the primary indicator of effectiveness lies in the low-perturbation regime, where AF++ demonstrates clear and consistent dominance.

\noindent\textbf{Qualitative Analysis on Semantic Information.}
To further visualize the semantic information encoded in VFs and AFs, we perform a cross-instance feature swapping experiment. Given two clean images $\mathbf{x}_1$ and $\mathbf{x}_2$, we extract their disentangled features $\{\mathbf{z}_{v}, \mathbf{z}_{a}\}$ using the pre-trained clean autoencoder and synthesize mixed images by swapping feature components:
$\tilde{\mathbf{x}}_{v1a2} = \mathcal{D}(\mathcal{HDF}(\mathbf{M}(\mathbf{z}_{v1}||\mathbf{z}_{a2})))$ and
$\tilde{\mathbf{x}}_{v2a1} = \mathcal{D}(\mathcal{HDF}(\mathbf{M}(\mathbf{z}_{v2}||\mathbf{z}_{a1})))$. As shown in Fig.~\ref{fig:featurevis}, the visual appearance of the reconstructed images is predominantly governed by the VF. Specifically, $\tilde{\mathbf{x}}_{v1a2}$ preserves the structural identity and texture of $\mathbf{x}_1$ regardless of the injected foreign AF $\mathbf{z}_{a2}$, confirming that our framework effectively isolates benign visual semantics within the VF. This is further supported by the progressive fusion results in Fig.~\ref{fig:featurevis_detail}: linearly interpolating the VF transitions the visual content from one object (e.g., bee) to another (e.g., bed), whereas varying the AF yields almost no perceptible visual changes.

Furthermore, the role of AFs becomes more evident through an analysis of the residual maps between the self-reconstructed image and the swapped image (e.g., $\tilde{\mathbf{x}}_1 - \tilde{\mathbf{x}}_{v1a2}$). As shown in the last two columns of Fig.~\ref{fig:featurevis}, these residuals are not random noise; rather, they emerge as spatially localized activation maps that closely align with foreground objects. For example, in the first row, the residual concentrates precisely on the ``black stork'', while in the second row, it highlights the ``dining table''. This phenomenon suggests that the AF captures the classifier’s attention regions and encodes high-level, class-discriminative cues that are largely independent of background texture. Consequently, when the original AF is replaced with a foreign one, the primary changes occur in these sensitive semantic regions, effectively ``rewriting'' the features used for classification without altering the human-perceptible structure of the image.

\begin{table*}[t!]
\small
\setlength{\tabcolsep}{0.03cm}
  \centering
  \caption{{Ablation study on the individual and combined effects of CD, HDF, and TI under fixed-target (label 864) and untargeted attacks. The setting without CD, HDF, or TI corresponds to the baseline DifAttack.}}
    \begin{tabular}{c|c|c|c|ccc|ccc|ccc|ccc|ccc}
    \hline\hline
    \multirow{2}[1]{*}{Settings} & \multirow{2}[1]{*}{CD} & \multirow{2}[1]{*}{HDF}& \multirow{2}[1]{*}{TI} & \multicolumn{3}{c|}{EfficientNet-B3} & \multicolumn{3}{c|}{ResNet-101} & \multicolumn{3}{c|}{ConvNeXtBase} & \multicolumn{3}{c|}{Swin-V2-T} & \multicolumn{3}{c}{Average} \\
\cline{5-19}          &       &       & &ASR   & Avg.Q & Med.Q & ASR   & Avg.Q & Med.Q & ASR   & Avg.Q & Med.Q & ASR   & Avg.Q & Med.Q & ASR   & Avg.Q & Med.Q \\
    \hline
    \multirow{4}[2]{*}{Targeted} &    -   &-&-       & 69.0    &                 6,080  &                 5,709  &                    62.5  &                 6,680  &                 6,954  & 97.0    &                 3,850  &                 3,461  &                    92.0  &                 4,291  &                 3,729  &                    80.1  &            5,225  &            4,963  \\

    & - & - & \checkmark 
& \multicolumn{3}{c|}{27.0\,{\scriptsize\textcolor{gray}{(25.5)}}\, - \, -} 
& \multicolumn{3}{c|}{31.5\,{\scriptsize\textcolor{gray}{(56.6)}}\, - \, -}   
& \multicolumn{3}{c|}{32.0\,{\scriptsize\textcolor{gray}{(45.0)}}\, - \, -}   
&  \multicolumn{3}{c|}{36.5\,{\scriptsize\textcolor{gray}{(52.0)}}\, - \, -} 
& \multicolumn{3}{c}{31.8\,{\scriptsize\textcolor{gray}{(47.3)}}\, - \, -}  \\

          &    -   & \checkmark  &-   & 70.0    &                 6,420  &                 6,496  &                    65.5  &                 6,582  &                 7,316  & 97.0    &                 2,908  &                 2,411  &                    96.5  &                 3,808  &                 3,016  &                    82.3  &            4,930  &            4,810  \\
        & \checkmark     &-     & - & 86.0    & \underline{   4,474 } & \underline{   3,416 } &                    73.5  & \underline{                5,715 } & \underline{                5,221 } & \textbf{100.0} &                 \underline{1,669}  &                 \underline{1,411}  &                    \underline{                   99.0 }  &                 2,632  & \underline{                2,116 } &                    89.6  & \underline{           3,623 } & \underline{           3,041 } \\
  & \checkmark     & \checkmark  & -  & \underline{89.0} &    4,818  &    4,224  & \underline{    75.5 } &    5,946  &    5,641  & \textbf{100.0} & {                1,706 } & {                1,456 } & \underline{                   99.0 } & \underline{   2,598 } &    2,176  & \underline{                   90.9 } &            3,767  &            3,374  \\
           & \checkmark     & \checkmark  & \checkmark  & \textbf{98.5} &    \textbf{1,287}  &    \textbf{766}  & \textbf{    94.0 } &    \textbf{1,460}  &    \textbf{226}  & \textbf{100.0} & \textbf{                710 } & \textbf{                301 } & \textbf{                   100.0 } & \textbf{   294 } &    \textbf{91}  & \textbf{                   98.1 } &            \textbf{939}  &            \textbf{346}  \\
           
    \hline \hline
    \multirow{4}[2]{*}{Untargeted} &  -     &  -   &-  & 100.0   &                      400  &                      151  &                    99.0  &                      523  &                      149  & 100.0   &                      435  &                      231  &                 100.0  &                      409  &                      171  &                    99.8  &                 442  &                 176  \\

       &  -     & -  &  \checkmark & \multicolumn{3}{c|}{73.5\,{\scriptsize\textcolor{gray}{(83.5)}}\, - \, -}     &                   \multicolumn{3}{c|}{73.0\,{\scriptsize\textcolor{gray}{(82.0)}}\, - \, -}   & \multicolumn{3}{c|}{73.0\,{\scriptsize\textcolor{gray}{(83.0)}}\, - \, -}    &                    \multicolumn{3}{c|}{73.5\,{\scriptsize\textcolor{gray}{(83.0)}}\, - \, -}   &                    \multicolumn{3}{c}{73.3\,{\scriptsize\textcolor{gray}{(82.9)}}\, - \, -} \\
       
         &   -    & \checkmark   &-  & 100.0   &                      865  &                      398  &     99.0  &      502  &      165  & 100.0   &                      534  &                      221  &                 100.0  &                      627  &                      255  &                    99.8  &                 632  &                 260  \\
         & \checkmark     & -     &- & 100.0   &      317  & \underline{      99 } &                    99.0  & \underline{                     434 } & \underline{                        80 } & 100.0   &                      189  &                         11  &                 100.0  &                      191  &                         64  &                    99.8  &                 283  & \underline{                   63 } \\
         & \checkmark     & \checkmark  &-   & 100.0   & \underline{                     302 } &                      129  & \textbf{    99.5 } &      482  &      161  & 100.0   & \underline{                     149 } & \underline{                            1 } &                 100.0  & \underline{                     170 } & \underline{                        57 } & \textbf{                   99.9 } & \underline{                276 } &                    87  \\
 & \checkmark     & \checkmark  & \checkmark  & {100.0} &    \textbf{138}  &    \textbf{1}  & \textbf{    99.5 } &    \textbf{172}  &    \textbf{1}  & {100.0} & \textbf{                83 } & \textbf{                1 } & {                   100.0 } & \textbf{   60 } &    \textbf{1}  & \textbf{                   99.9 } &            \textbf{113}  &            \textbf{1}  \\
    \hline\hline
    \end{tabular}%
    \vspace{1mm}
\begin{minipage}{0.98\linewidth}
\footnotesize
\textit{Note:} The row with CD=--, HDF=--, and TI=\checkmark reports the one-query initial ASR of autoencoder-reconstructed transferable AEs, rather than DifAttack equipped with TI and iterative optimization. Gray numbers in parentheses indicate the ASR of the original transferable AEs before autoencoder reconstruction. Since this row only evaluates the initialization itself with one query, Avg.Q and Med.Q are denoted by ``--''. 
\end{minipage}
  \label{tab:ablationArch}%
\end{table*}%

\subsection{Ablation Studies}
\label{sec:exp:ablation}

To substantiate the contribution of each proposed component, we conduct the ablation study starting from the baseline (original DifAttack). We investigate three key enhancements: 1) the Cross-Domain \textbf{(CD)} training scheme, 2) the Hierarchical Decouple-Fusion \textbf{(HDF)} module, and 3) the Transferable Initialization \textbf{(TI)} strategy. The quantitative results are summarized in Tables \ref{tab:ablationArch} and~\ref{tab:C2Ablation}.

\noindent\textbf{CD.} Replacing the single shared autoencoder with the cross-domain architecture yields significant performance gains.  {As shown in the CD-only rows of Table \ref{tab:ablationArch}, CD consistently improves the targeted ASR on all four complex victim models. It also improves the overall targeted and untargeted query efficiency. Notably, in the challenging targeted attacks, CD boosts the average ASR by $9.5\%$ and reduces the average Avg.Q by $30.7\%$}, providing a solid foundation for the framework.

\noindent\textbf{HDF.}
{The transition from the single-layer DF module to the two-stage HDF module provides a useful hierarchical refinement of the AF/VF decomposition. As shown in Table~\ref{tab:ablationArch}, under the fixed-target setting, adding HDF alone to the DifAttack baseline improves both targeted ASR and query efficiency, indicating that HDF can improve adversarial-feature optimization without CD. When combined with CD, its benefit remains clear, especially when the ASR has not yet saturated. Under the easier second-highest-score target setting in Table~\ref{tab:C2Ablation}, CD+HDF achieves the best overall performance among all variants without TI. These results show that HDF and CD are complementary: CD provides more stable cross-domain AF/VF separation, while HDF further refines the preliminary disentangled features before fusion. The effect of HDF can also depend on target difficulty and ASR saturation. When ASR is already saturated or near-saturated, HDF may preserve ASR with slight query fluctuations or marginal ASR changes. Similarly, in the untargeted setting, HDF alone preserves ASR but may slightly increase query counts on some saturated models, while CD+HDF provides a stronger overall trade-off by reducing query costs and sometimes further improving ASR. Overall, these ablation results show that HDF is an effective hierarchical refinement module, especially when combined with CD.}

\noindent\textbf{TI.} {Finally, TI provides a transferable initialization for the subsequent query-based optimization. However, TI alone is still far from the final performance of our full method, especially in the targeted setting. As shown in Table \ref{tab:ablationArch}, under the fixed-target setting, the autoencoder-reconstructed TI only achieves an average ASR of $31.8\%$, whereas the full method with CD, HDF, and TI reaches $98.1\%$. Similarly, in the untargeted setting, TI alone obtains an average ASR of $73.3\%$, which is still much lower than the $99.9\%$ ASR achieved by the full method. Moreover, our attack does not directly start from the original transferable AEs. Instead, these samples are first reconstructed by the autoencoder and then used as the initialization, which leads to a lower ASR than the original transferable samples, as indicated by the gray numbers in Table \ref{tab:ablationArch}. This gap further suggests that the final performance is not mainly inherited from the transferable samples themselves. Rather, TI provides an effective warm start, and its benefit is further amplified when combined with the learned disentangled latent space and subsequent query-based optimization.}

\begin{table*}[t!]
  \centering
  \small
  \setlength{\tabcolsep}{0.05cm}
  \caption{{Ablation study under an easier targeted setting using the second-highest prediction-score class as the target.}}
    \begin{tabular}{c|c|c|ccc|ccc|ccc|ccc|ccc}
    \hline\hline
    \multirow{2}[1]{*}{CD} & \multirow{2}[1]{*}{HDF} & \multirow{2}[1]{*}{TI} &\multicolumn{3}{c|}{EfficientNet-B3} & \multicolumn{3}{c|}{ResNet-101} & \multicolumn{3}{c|}{Swin-V2-T} & \multicolumn{3}{c|}{ConvNeXtBase} & \multicolumn{3}{c}{Average} \\
\cline{4-18}          &       && ASR   & Avg.Q & Med.Q & ASR   & Avg.Q & Med.Q & ASR   & Avg.Q & Med.Q & ASR   & Avg.Q & Med.Q & ASR   & Avg.Q & Med.Q \\
    \hline
        -& - &-       &     68.5  &    6,219  &    7,104  &                    57.5  &                 6,804  &                 8,626  &                    \underline{99.0}  &                 2,449  &                 2,056  &         \underline{99.5}  &       2,279  &       1,901  &                    81.1  &                 4,437  &                 4,922  \\

      -& - &\checkmark       &      \multicolumn{3}{c|}{63.5\,{\scriptsize\textcolor{gray}{(72.0)}}\, - \, -}  &                    \multicolumn{3}{c|}{36.5\,{\scriptsize\textcolor{gray}{(80.0)}}\, - \, -} &                    \multicolumn{3}{c|}{55.0\,{\scriptsize\textcolor{gray}{(78.0)}}\, - \, -} &         \multicolumn{3}{c|}{49.0\,{\scriptsize\textcolor{gray}{(72.5)}}\, - \, -}  &                    \multicolumn{3}{c}{51.0\,{\scriptsize\textcolor{gray}{(75.6)}}\, - \, -} \\
      
         - & \checkmark     & - &                   70.0  &    5,995  &    5,751  &                    63.5  &    6,346  &    6,864  &                    \underline{99.0}  &                 2,553  &                 2,036  &                    {99.0}  &                 2,677  &                 2,241  &                    82.9  &                 4,393  &                 4,223  \\
    \checkmark     & -      & - &       70.5  & \underline{                5,862 } &                 5,949  &                    61.5  &                 6,464  &                 7,134  &                    97.0  &                 2,756  &                 2,036  & \textbf{                100.0 } & \underline{                1,019 } &                      856  &                    82.3  &                 4,025  &                 3,994  \\
    \checkmark     & \checkmark     & -&\underline{                   72.5 } &                 5,965  & \underline{                5,746 } & \underline{                   70.5 } & \underline{   5,551 } & \underline{   4,734 } & \textbf{   100.0 } & \underline{   2,234 } & \underline{   1,896 } & \textbf{   100.0 } &    1,032  & \underline{     825 } & \underline{                   85.8 } & \underline{                3,696 } & \underline{                3,300 } \\
    \checkmark     & \checkmark     &  \checkmark &\textbf{                   100.0 } &                 458  & \textbf{                1 } & \textbf{                   94.0 } & \textbf{   1,320 } & \textbf{   204 } & \textbf{   100.0 } & \textbf{   209 } & \textbf{   1 } & \textbf{   100.0 } &    \textbf{485}  & \textbf{     53 } & \textbf{                   98.5 } & \textbf{                618 } & \textbf{                65 } \\
    \hline\hline
    \end{tabular}%
    \vspace{1mm}
\begin{minipage}{0.98\linewidth}
\footnotesize
\textit{Note:} The row with CD=--, HDF=--, and TI=\checkmark reports the one-query initial ASR of autoencoder-reconstructed transferable AEs, rather than DifAttack equipped with TI and iterative optimization. Gray numbers in parentheses indicate the ASR of the original transferable AEs before autoencoder reconstruction. Since this row only evaluates the initialization itself with one query, Avg.Q and Med.Q are denoted by ``--''.
\end{minipage}
  \label{tab:C2Ablation}%
\end{table*}%

\begin{figure*}[t]
\centering
\includegraphics[width=1\textwidth]{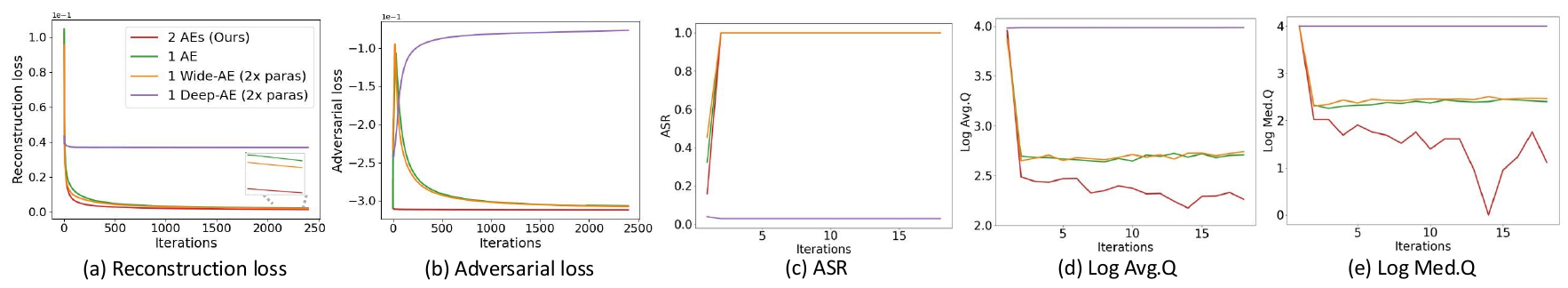} 
\caption{Comparison of training dynamics and attack performance under equivalent parameter budgets. The cross-domain architecture (Red) consistently outperforms parameter-matched baselines (Wide/Deep-AE) in minimizing reconstruction/adversarial losses (Left) and maximizing attack efficiency (Right). Note: The x-axis values differ slightly between loss and attack metrics due to different checkpoint recording intervals.
}
\label{fig:recloss}
\end{figure*}

\subsection{Impact of Cross-Domain Architecture}
\label{sec:exp:dual_domain}

A critical question is whether the performance gains of DifAttack++ stem from the proposed cross-domain disentanglement mechanism or simply from the increased model capacity (i.e., doubled parameters compared to the single-encoder baseline). To isolate the source of improvement, we conduct a comparative study focusing on model capacity and optimization dynamics.

Specifically, we construct two parameter-equivalent variants of the single-autoencoder baseline (denoted as \textbf{1 AE}, $\sim12.8$ MB): 1) \textbf{Deep-AE}: Increasing the network depth by stacking convolutional layers. 2) \textbf{Wide-AE}: Expanding the channel dimensions of the encoder/decoder. Both variants match the parameter budget ($\sim25.8$ MB) of our cross-domain framework (\textbf{2 AEs} without HDF module). We track two critical metrics during training to evaluate the quality of the learned representations: self-reconstruction Loss ($\mathcal{L}_{rec}$), which measures the capability of the autoencoders to faithfully reconstruct images within their respective domains, and the averaged adversarial loss $\mathcal{L}_{adv}$ of the feature-exchanged AEs $\tilde{\mathbf{x}}^* = \mathcal{D}^*(\mathcal{HDF}^*(\mathbf{M^*}(\mathbf{z}_v|| \mathbf{z}_a^*)))$.
The comparative trajectories are plotted in Fig. \ref{fig:recloss}. As observed in the reconstruction loss curve in Fig. \ref{fig:recloss}(a), the cross-domain architecture (red curve) achieves the fastest convergence and the lowest final loss compared to all single-autoencoder variants (Wide/Deep/1-AE). This indicates that by physically isolating clean and adversarial domains, our framework enables each encoder-decoder pair to specialize in its respective domain, leading to more accurate reconstruction without requiring excessive model depth or width.

Furthermore, the adversarial loss curve in Fig.~\ref{fig:recloss}(b) demonstrates the effectiveness of our disentanglement strategy. This metric quantifies how ``adversarial" the synthesized image appears when an AF is injected into a clean VF: the lower the value, the stronger the adversarial capability. Our method (red curve) starts from a significantly lower initial value and consistently outperforms the baselines, confirming that the disentangled representation successfully preserves adversarial semantics during synthesis.
This advantage also translates into improved query efficiency. As shown in Figs.~\ref{fig:recloss}(c–d), the cross-domain architecture achieves a substantially lower query cost from the early stages of optimization. In contrast, the Wide-AE performs comparably to the baseline, suggesting that merely increasing model capacity cannot alleviate the feature entanglement bottleneck. The Deep-AE exhibits suboptimal and unstable convergence, likely due to gradient vanishing or explosion in deeper networks lacking specialized designs such as residual connections.
Together, these results demonstrate that simply scaling up model capacity—via width (Wide-AE) or depth (Deep-AE)—yields only marginal or unstable improvements. The significant performance gain of DifAttack++ thus stems not from increased parameters, but from the structural advantage of cross-domain disentanglement.

\section{Conclusion}
\label{sec:con}
This work presents a novel query-efficient score-based black-box adversarial attack method named DifAttack++, which hierarchically disentangles an image's latent representation into AFs and VFs and performs black-box attacks in the learned disentangled feature space. The main idea is to optimize the AF while keeping the VF unchanged until a successful AE is obtained. Experimental results demonstrate the superior attack efficiency of DifAttack++ in both closed-set and open-set scenarios, while maintaining satisfactory visual quality of the generated AEs. {Future work may further improve the robustness of the learned AF space by incorporating more diverse surrogate ensembles or flat-local-maxima optimization objectives during autoencoder training.}

\section*{Acknowledgments}
{The authors would like to thank Li Zheng from the University of Macau for a helpful comment related to attack initialization. We are grateful to the anonymous reviewers for their valuable guidance and insightful comments. J. Zhou's, J. Zeng's, J. Tian's, and part of J. Liu's work was supported in part by Macau Science and Technology Development Fund under 001/2024/SKL, 0119/2024/RIB2 and 0110/2025/R1B2; in part by Research Committee at University of Macau under MYRG-CRG2025-00031-FST and MYRG-GRG2025-00086-FST; in part by the Guangdong Basic and Applied Basic Research Foundation under Grant 2024A1515012536. I. Echizen's and part of J. Liu's work was supported by JSPS KAKENHI Grant JP24H00732; by JST CREST Grants JPMJCR20D3 and JPMJCR2562 including AIP challenge program; and by JST K Program Grant JPMJKP24C2 Japan.}

\bibliographystyle{IEEEtran_nodash}
\bibliography{ref}


 

\begin{IEEEbiography}[{\includegraphics[width=1in,height=1.25in,clip,keepaspectratio]{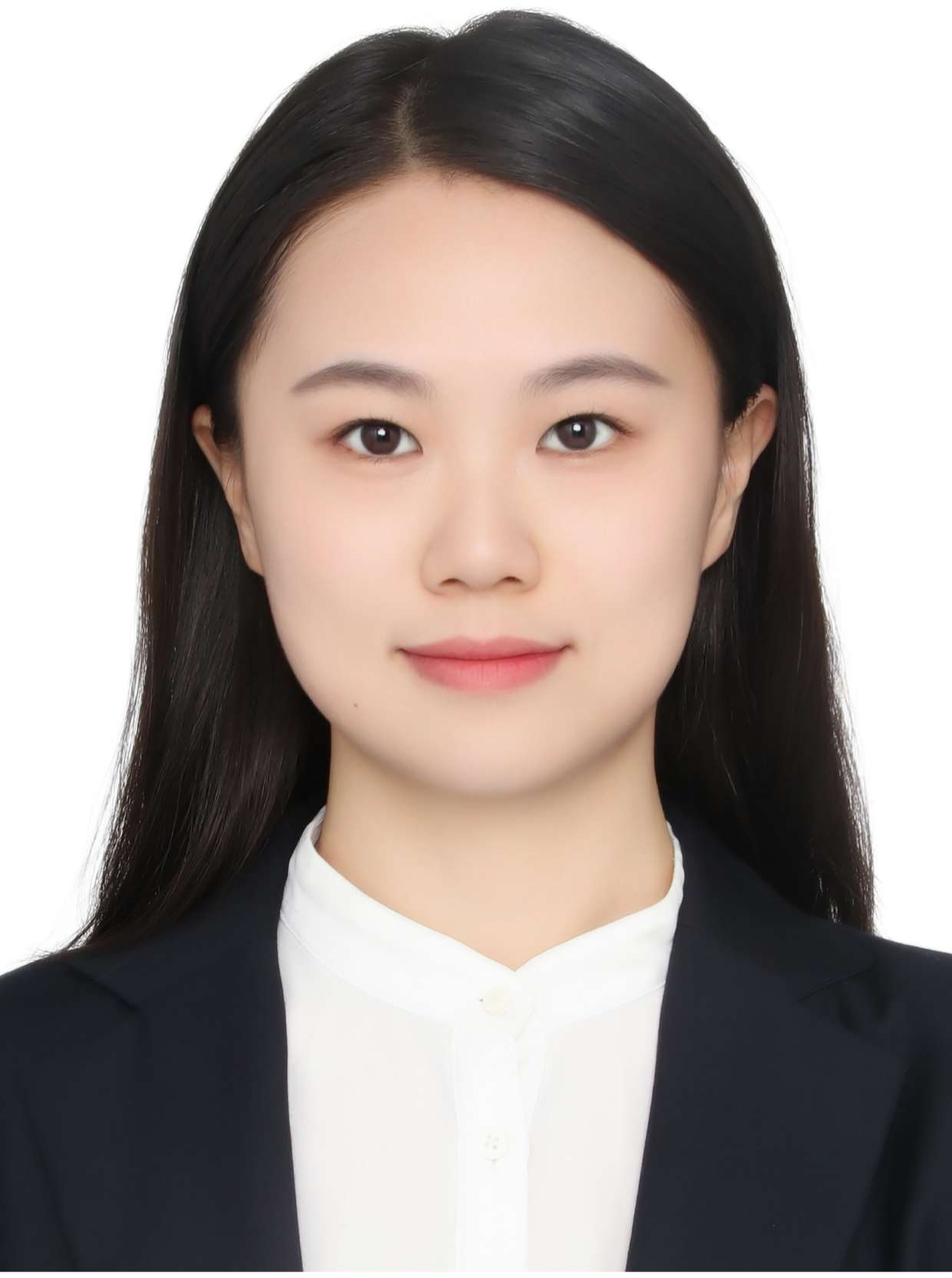}}]{Jun Liu} (Member, IEEE) is currently a Postdoctoral Researcher at the National Institute of Informatics, Japan. She received the B.Eng. and M.Eng. degrees from the Department of Software Engineering, Beihang University, China, in 2015 and 2018, respectively, and the Ph.D. degree from the Department of Computer and Information Science, University of Macau, China, in 2025. She previously worked as a machine learning algorithm engineer at Tencent, Shenzhen, China. Her research interests include adversarial machine learning, privacy protection, image forensics, and multimedia signal processing.

\end{IEEEbiography}

\begin{IEEEbiography}
[{\includegraphics[width=1in,height=1.25in,clip]{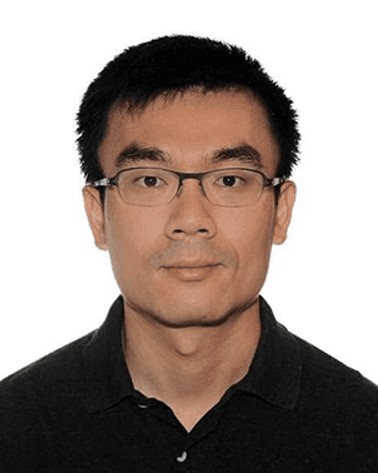}}]
{Jiantao Zhou} (Senior Member, IEEE) received the B.Eng. degree from the Department of Electronic Engineering, Dalian University of Technology, in 2002, the M.Phil. degree from the Department of Radio Engineering, Southeast University, in 2005, and the Ph.D. degree from the Department of Electronic and Computer Engineering, Hong Kong University of Science and Technology, in 2009. He held various research positions with the University of Illinois at Urbana-Champaign, Hong Kong University of Science and Technology, and McMaster University. He is currently a Professor with the Department of Computer and Information Science, Faculty of Science and Technology, and Director of the Research Services and Knowledge Transfer Office, University of Macau. His research interests include multimedia security and forensics, multimedia signal processing, artificial intelligence, and big data. He holds four granted U.S. patents and two granted Chinese patents. He has coauthored two papers that received the Best Paper Award at the IEEE Pacific-Rim Conference on Multimedia in 2007 and the Best Student Paper Award at the IEEE International Conference on Multimedia and Expo in 2016. He is serving as a Senior Editor for the IEEE Transactions on Dependable and Secure Computing, and as an Associate Editor for the IEEE Transactions on Image Processing and the IEEE Transactions on Multimedia. 
\end{IEEEbiography}

\begin{IEEEbiography}
[{\includegraphics[width=1in,height=1.25in,clip]{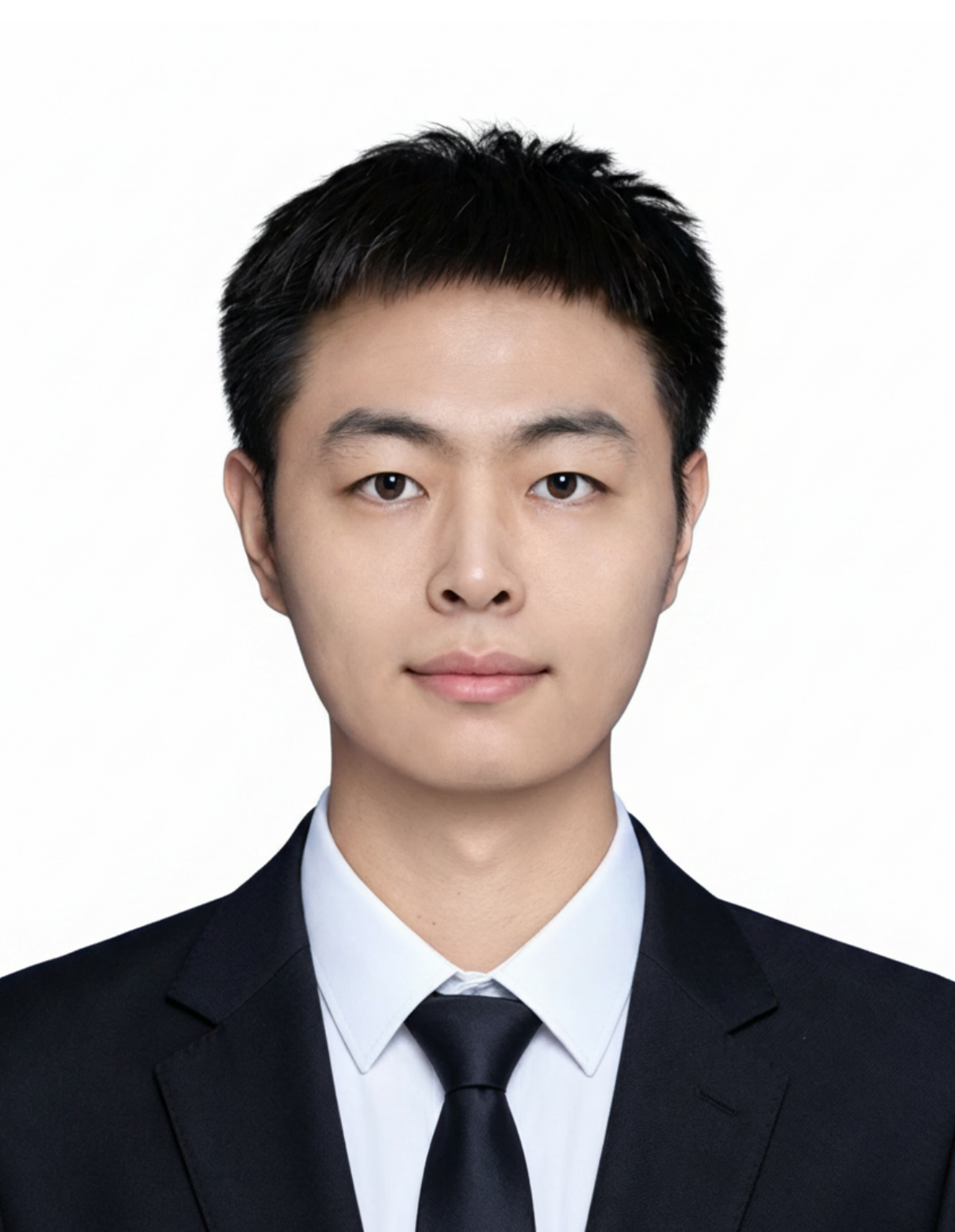}}]
{Jiandian Zeng} (Member, IEEE) received the Ph.D. degree in Department of Computer and Information Science, University of Macau, Macau, China, in 2023. He is currently a lecturer at the Advanced Institute of Natural Sciences, Beijing Normal University. His research interests include edge intelligence and knowledge engineering.
\end{IEEEbiography}

\begin{IEEEbiography}
[{\includegraphics[width=1in,height=1.25in,clip]{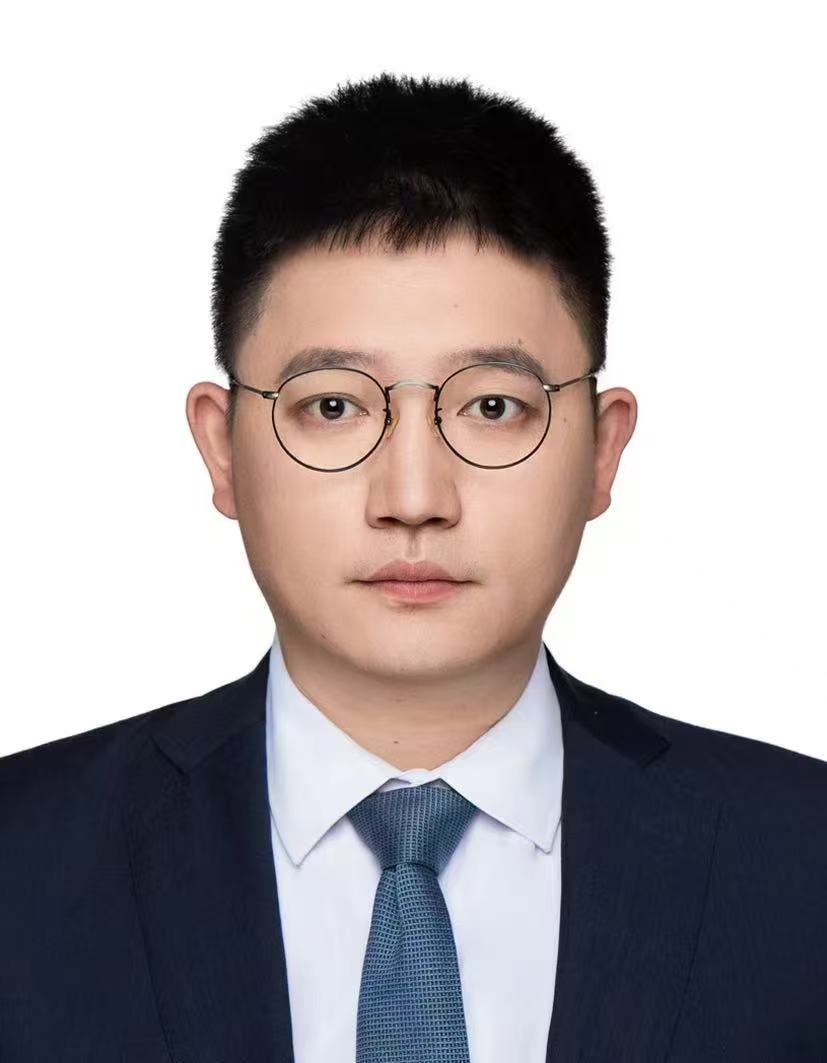}}]
{Jinyu Tian} (Member, IEEE) received the B.S.
and M.S. degrees in mathematics from Chongqing
University, Chongqing, China, in 2014 and 2017,
respectively, and the Ph.D. degree in computer science
from the University of Macau, Macau, China,
in 2022. He is currently an Assistant Professor
at the School of Computer Science and Engineering,
Macau University of Science and Technology. His current research interests include adversarial machine learning, security of
deep learning, multimedia forensics, and subspace learning. He is an active
reviewer for numerous prestigious journals and conferences, such as IEEE
Transactions on Image Processing, IEEE Transactions on Multimedia,
ACM MM, CVPR, NeurIPS, and ICML.
\end{IEEEbiography}

\begin{IEEEbiography}[{\includegraphics[width=1in,height=1.25in,clip,keepaspectratio]{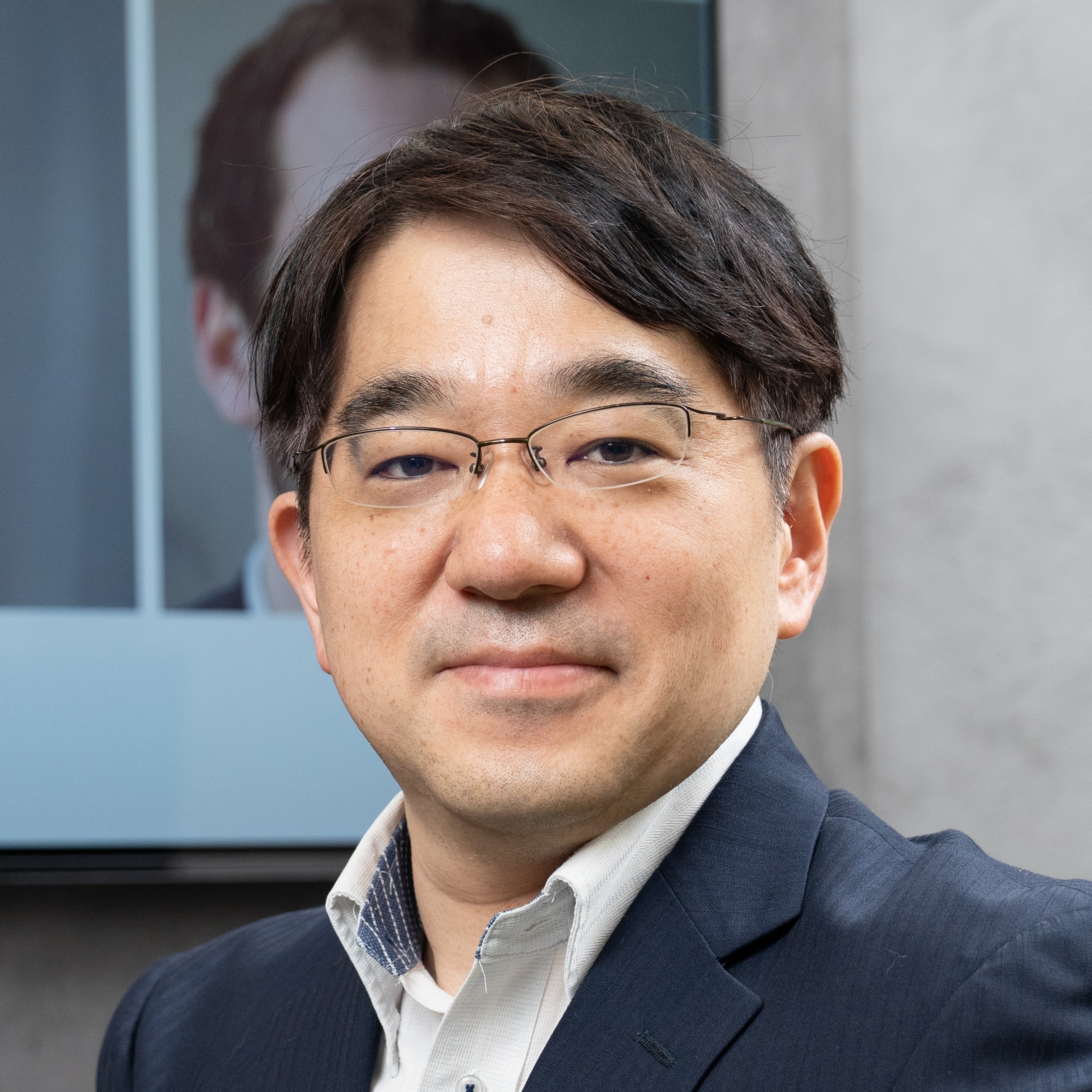}}]{Isao Echizen} (Senior Member, IEEE) received B.S., M.S., and D.E. degrees from the Tokyo Institute of Technology, Japan, in 1995, 1997, and 2003, respectively. He joined Hitachi, Ltd. in 1997, where he worked as a research engineer in the company's Systems Development Laboratory until 2007. He is currently a director and professor in the Information and Society Research Division, as well as the director of the Global Research Center for Synthetic Media at the National Institute of Informatics (NII), Japan. He also holds concurrent professorships at the Graduate School of Information Science and Technology, the University of Tokyo, and the Graduate Institute for Advanced Studies, SOKENDAI (The Graduate University for Advanced Studies). He has served as a visiting professor at the University of Freiburg and the University of Halle-Wittenberg in Germany.
His current research interests include AI security, multimedia security, and multimedia forensics. From 2020 to 2026, he served as a research director for the CREST FakeMedia project funded by the Japan Science and Technology Agency (JST). He is currently a research director for the K Program SYNTHETIQ X project and a co-research director for the CREST Social AI project, both supported by JST.
He is a Fellow of the IEICE and the IPSJ, and a Senior Member of the IEEE, where he previously served as a member of the Information Forensics and Security Technical Committee of the IEEE Signal Processing Society. He is also the Japanese representative to IFIP and IFIP TC11 (Security and Privacy Protection in Information Processing Systems), a vice president of APSIPA, and an editorial board member for several journals, including IEEE Transactions on Dependable and Secure Computing, EURASIP Journal on Image and Video Processing, and Elsevier's Journal of Information Security and Applications.
He has received numerous awards, including the Ichimura Prize in Science for Excellent Achievement (2026), the Commendation for Science and Technology by the Minister of Education, Culture, Sports, Science and Technology (Research Category, 2025), the IEICE Achievement Award (2025), the IEEE BTAS/IJCB 5-Year Highest Impact Award (2024, 2023), the IEICE Best Paper Award (2023), the IPSJ Best Paper Awards (2005, 2014), the IPSJ Nagao Special Researcher Award (2011), the DOCOMO Mobile Science Award (2014), the Information Security Cultural Award (2016), and the IEEE WIFS Best Paper Award (2017).
\end{IEEEbiography}


\end{document}